\documentclass[10pt,twocolumn,letterpaper]{article}

\usepackage{cvpr}

\usepackage{multirow}
\usepackage{algorithm}
\usepackage{algpseudocode}
\usepackage{caption}

\definecolor{cvprblue}{rgb}{0.21,0.49,0.74}
\definecolor{Red}{RGB}{192,0,0}
\definecolor{Blue}{RGB}{12,114,186}
\definecolor{GradStart}{HTML}{7A28CB}
\definecolor{GradEnd}{HTML}{00D2FF}

\newcommand{\LiveLight}{%
\textcolor{GradStart!100!GradEnd}{L}%
\textcolor{GradStart!87!GradEnd}{i}%
\textcolor{GradStart!75!GradEnd}{v}%
\textcolor{GradStart!62!GradEnd}{e}%
\textcolor{GradStart!50!GradEnd}{L}%
\textcolor{GradStart!37!GradEnd}{i}%
\textcolor{GradStart!25!GradEnd}{g}%
\textcolor{GradStart!12!GradEnd}{h}%
\textcolor{GradStart!0!GradEnd}{t}%
}

\usepackage[
    pagebackref,
    breaklinks,
    colorlinks,
    allcolors=cvprblue
]{hyperref}

\def\paperID{*****}
\def\confName{CVPR}
\def\confYear{2026}

\title{{\LiveLight}: Real-time Streaming Video Relighting with Interactive Control}

\author{
Yue Ma$^{1}$ \quad
Jiangming Wang$^{1}$ \quad
Yucheng Wang$^{1}$ \quad
Xilai Wang$^{1}$ \quad
Zhiyuan Li$^{2}$\\
Xinyu Wang$^{3}$ \quad
Hongyu Liu$^{1}$ \quad
Ruofan Liang$^{4}$ \quad
Songchun Zhang$^{1}$ \quad
Yuxuan Xue$^{5,\ddagger}$ \quad
Qifeng Chen$^{1,\dagger}$\\[6pt]
$^{1}$HKUST \quad
$^{2}$University of Macau \quad
$^{3}$THU \quad
$^{4}$UoT \quad
$^{5}$University of Tuebingen\\[6pt]
Project Page: \url{https://living-lighting.github.io}
}

\begin{document}

\twocolumn[{
\renewcommand\twocolumn[1][]{#1}

\maketitle

\vspace{-32pt}

\begin{center}
    \centering
    \includegraphics[width=\linewidth]{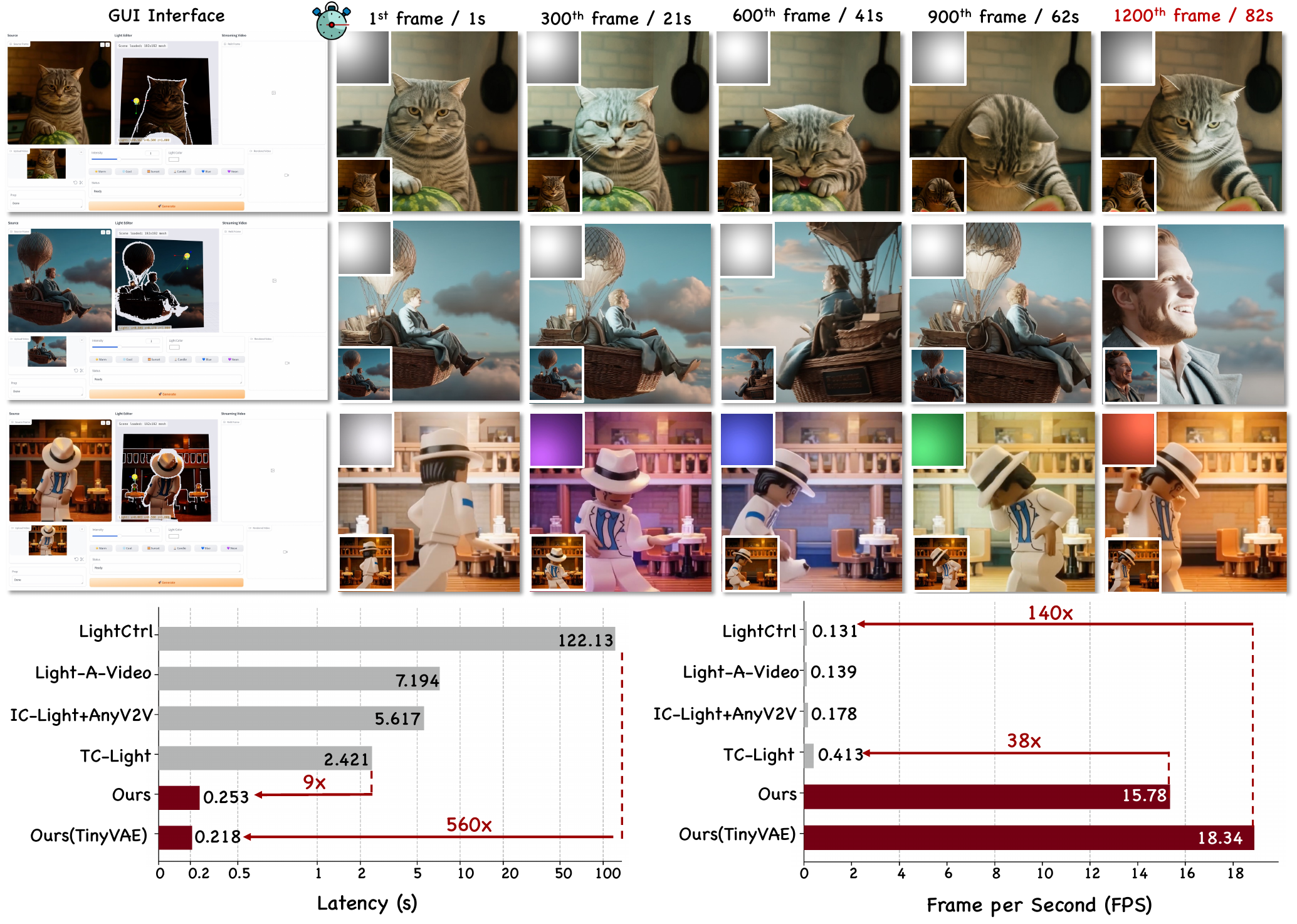}
    \captionof{figure}{
        \textbf{Showcase of the proposed LiveLight.}
        LiveLight enables real-time streaming video relighting with
        interactive 3D point-light control, supporting dynamic adjustment
        of light position, intensity, and color while preserving temporal
        coherence and visual quality.
    }
    \label{fig:teaser}
\end{center}
}]

\begingroup
\renewcommand{\thefootnote}{}
\footnotetext{%
    \textsuperscript{\ensuremath{\dagger}} Corresponding author.
    \qquad
    \textsuperscript{\ensuremath{\ddagger}} Project leader.
}
\endgroup

\begin{abstract}
We present LiveLight, the first diffusion-based framework for real-time streaming video relighting with interactive 3D lighting control. Achieving this is non-trivial, as it requires overcoming three critical challenges: effectively injecting dynamic 3D lighting into a diffusion model, maintaining high-fidelity generation under an extremely low NFE (Number of Function Evaluations) budget for real-time speed, and facilitating continuous streaming for interactive control. To address these pain points, we propose three key designs. First, for accurate lighting injection, we propose a lightweight adapter that feeds Multi-Plane Light Irradiance (MPLI) conditions—depth-aware irradiance maps encoding 3D lighting geometry—directly into the diffusion backbone. Second, to prevent rendering quality degradation at low NFEs towards real-time distillation, we introduce a geometry-guided feedback branch. This training-time constraint leverages a frozen geometry estimator to enforce depth- and normal-consistent relighting, ensuring geometrically plausible shading without adding inference overhead. Finally, to enable streaming interaction, we develop a progressive rolling-window strategy that maintains a denoising ladder of latent chunks at varying noise levels. By propagating intermediate states, this strategy guarantees temporal coherence and supports arbitrarily long video relighting with per-frame reference refresh. Extensive experiments on real-world and synthetic benchmarks demonstrate that LiveLight achieves state-of-the-art relighting quality while running at real-time speed, significantly outperforming offline baselines in temporal stability, lighting controllability, and user preference. To foster real-time interactive relighting research, we will publicly release our models, training data, and synthetic data generator.
\end{abstract}
\section{Introduction}
\label{sec:intro}

Video relighting is a fundamental capability for film production, virtual cinematography, and live-streaming content creation, enabling the alteration of illumination in captured scenes. In these demanding workflows, real-time streaming video relighting introduces a transformative, interactive paradigm. By responding to continuously changing light parameters with minimal latency and ensuring strict temporal coherence across consecutive frames, this capability unlocks unprecedented creative freedom. For instance, a user can dynamically drag a virtual light source through a scene and instantly preview how shadows, highlights, and colors evolve over a video stream, entirely bypassing the prohibitive delays of offline re-rendering. Ultimately, this seamless integration of high-fidelity relighting and real-time interaction redefines the boundaries of dynamic content creation, bridging the gap between professional studio effects and accessible, on-the-fly editing.

Recent diffusion-based relighting methods have made impressive progress on individual aspects of this problem.
For image relighting, IC-Light~\cite{zhang2025scaling} fine-tunes a pretrained diffusion model for light-conditioned generation and achieves strong results. For video relighting,  LightCtrl~\cite{peng2026lightctrl} introduces lightweight physical priors to improve controllability without dense intrinsic supervision.
Light-A-Video~\cite{zhou2025light} and TC-Light~\cite{liu2026tc} combine per-frame image relighting with video diffusion priors in a training-free manner, and RelightVid~\cite{fang2025relightvid} trains temporal models on curated relighting datasets.

\begin{figure}  
    \centering
    \includegraphics[width=\linewidth]{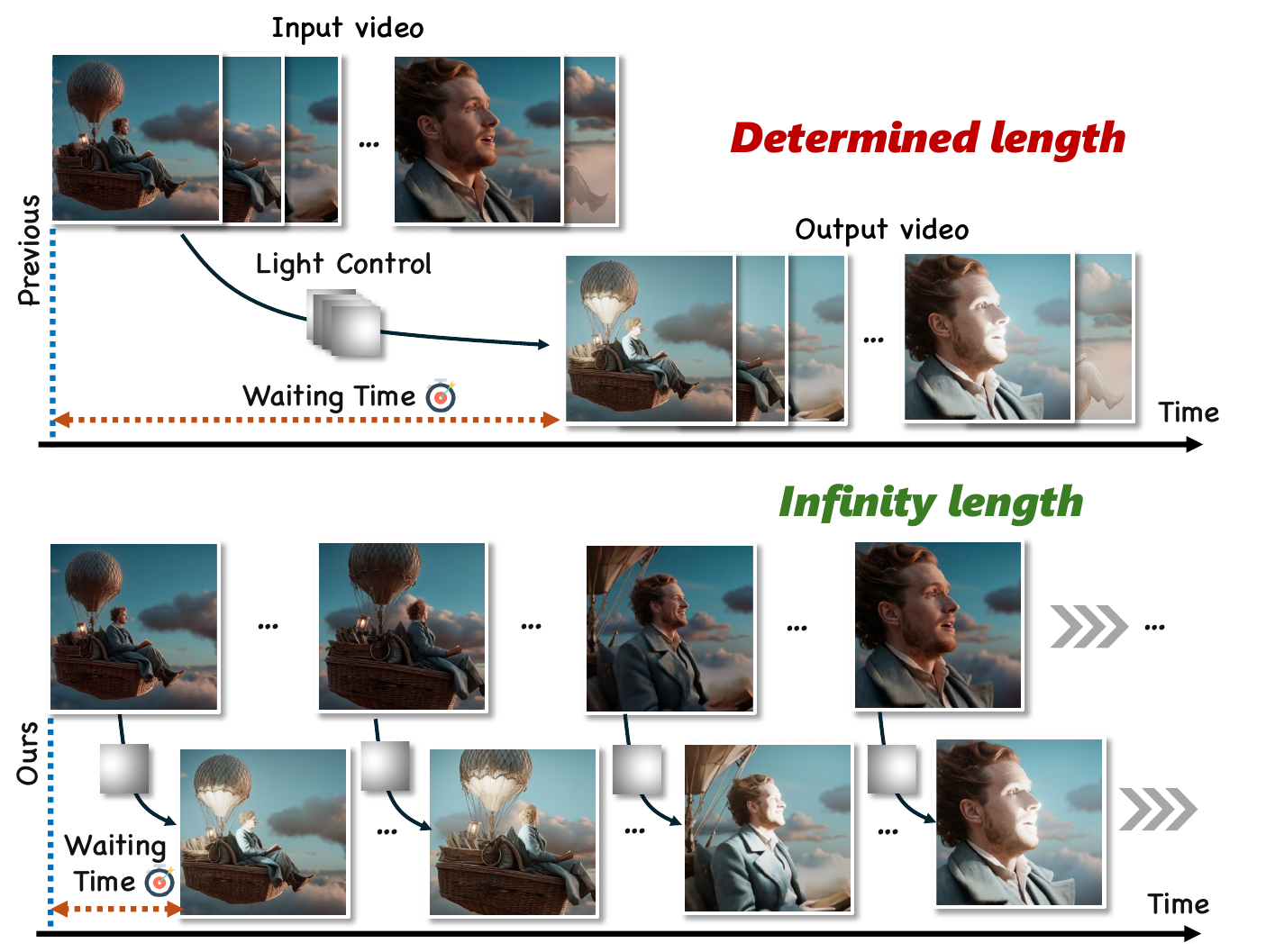}
    \caption{\textbf{Motivation for LiveLight.} Previous offline video relighting methods require the entire input video and a pre-defined lighting trajectory before generation, producing a fixed-length output after a long waiting time. In contrast, LiveLight processes input frames in a streaming fashion with per-frame dynamic lighting control, producing relit frames with low latency and supporting arbitrarily long video generation.}
    \label{fig:motivation}
\end{figure}

Despite these advances, enabling real-time interactive video relighting remains highly challenging. Existing video relighting methods operate in an offline, clip-level paradigm: the full lighting trajectory must be specified before generation, and the entire video is produced in a single pass. This design precludes real-time interactive control and imposes a high computational cost that scales linearly with video length. Alternatively, applying image-based relighting models independently to each frame can process inputs incrementally, but this inevitably produces severe temporal flicker, rendering it unsuitable for coherent video streaming.
%


An ideal real-time video relighting system should be interactive, streaming, and high-fidelity, capable of providing instant visual feedback as the user adjusts the lighting direction. Intuitively, achieving this requires a fundamental departure from the conventional clip-level generation paradigm. Instead, the generation process must be fine-grained to the frame level and operate under an extremely low NFE (Number of Function Evaluations) budget to guarantee real-time latency. Furthermore, it necessitates a genuine streaming mechanism to ensure that the continuously generated frames remain strictly temporally coherent, as illustrated in Fig.~\ref{fig:motivation}. Driven by these rigorous requirements, we present \textbf{LiveLight}, the first diffusion-based framework tailored to solve these challenges. To this end, we introduce three complementary designs:

(i) \textbf{\textit{Lightweight adapter for interactive light injection.}}
To support fine-grained, frame-level interactive relighting, we use the Multi-Plane Light Irradiance (MPLI) representation~\cite{bian2025relightmaster} as a compact geometry-aware carrier of user-specified lighting parameters. The MPLI maps are then encoded by a lightweight adapter with only \(\sim 1\)M parameters into compact lighting tokens, which interact with the diffusion features through cross-attention in shallow layers. This localized injection strategy introduces spatially varying illumination cues at an early stage, while avoiding unnecessary perturbation to deeper denoising blocks that encode appearance priors. As a result, the model preserves identity and texture fidelity with negligible overhead, allowing changes in light position, intensity, and color to be reflected immediately for interactive control.

(ii) \textbf{\textit{Real-time distillation with geometry-guided feedback.}}
To achieve real-time interactive relighting, we do not simply truncate the
diffusion sampling process at inference time. Instead, we distill the model
with the same compact few-step rollout used during testing, enabling it to
produce effective illumination updates under an extremely low NFE budget.
However, aggressive step reduction can weaken the geometric consistency of
synthesized highlights, shadows, and depth-dependent illumination. To address
this issue, we introduce a training-time geometry-guided feedback branch. A
frozen geometry estimator predicts depth and normal cues from the generated
image and back-propagates geometry-aware gradients to the relighting model.
This supervision encourages the generated shading to remain consistent with
the underlying structure, compensating for the quality degradation
caused by a few-step inference without introducing any inference-time cost.

(iii) \textbf{\textit{Rolling-Window Streaming Video Relighting.}}
To fundamentally break away from the conventional offline, clip-level paradigm, we introduce a genuine streaming mechanism formulated as a rolling-window process over micro-chunks. A sliding window of latent micro-chunks is maintained, where at each streaming step the video denoising backbone jointly processes all chunks: the oldest chunk is fully denoised, decoded, and emitted, while the remaining chunks are refined as temporal context and a new noisy chunk is appended to the tail. This design achieves strict temporal coherence through persistent latent propagation without redundant overlap blending. Crucially, it refreshes the reference and lighting conditions per frame, keeping appearance, geometry, and lighting synchronized with the input video to support dynamic, arbitrary-length interactive sessions.

Extensive experiments demonstrate that LiveLight achieves state-of-the-art relighting quality on both real-world and synthetic benchmarks while running at real-time speed. Compared with offline methods, LiveLight produces more temporally stable results over long videos and enables interactive lighting adjustment with low latency. A user study further confirms that our method is preferred over existing baselines in lighting accuracy, appearance consistency, and temporal coherence.

In summary, our contributions are as follows:
\begin{itemize}
    \item We propose LiveLight, the first diffusion-based framework for real-time, streaming video relighting with interactive lighting control.
    \item We design a lightweight adapter that efficiently integrates the MPLI lighting representation into the diffusion backbone, and introduce a geometry-guided feedback branch that enforces structure-preserving shading at training time without inference overhead.
    \item We design a rolling-window generation strategy with per-frame reference refresh that propagates intermediate latent states for temporal coherence, enabling low-latency and arbitrarily long video relighting.
    \item Extensive experiments on real-world and synthetic benchmarks show that LiveLight achieves superior quality, efficiency, and user preference over state-of-the-art methods.
\end{itemize}

\section{Related Work}
\subsection{Relighting Method}
Relighting has been extensively studied in graphics, where traditional approaches rely on inverse rendering~\cite{chen2025contextflow, debevec2008rendering, ma2025controllable, zhang2021physg, zhang2022modeling} or light stage captures~\cite{debevec2000acquiring} to decompose and re-render scenes under novel illumination. While physically grounded, these methods require specialized inputs and are largely restricted to portraits~\cite{pandey2021total}. Recent diffusion-based methods~\cite{zhang2025scaling, zeng2024dilightnet, ma2024followyouremoji, ma2025followyourmotion, ma2026fastvmt, liu2026opsd, ma2025followcreation, wang2026liveedit, kim2024switchlight, jin2024neural, chaturvedi2025synthlight, ren2024relightful, kocsis2024intrinsic} have demonstrated flexible image relighting with diverse control modalities, among which IC-Light~\cite{zhang2025scaling} has become a widely adopted backbone owing to its effective light transport formulation. Extending to video, RelightVid~\cite{fang2025relightvid} trains temporal models on curated relighting datasets, while Light-A-Video~\cite{zhou2025light} and LightCtrl~\cite{peng2026lightctrl} adopt training-free pipelines that combine per-frame image relighting with video diffusion priors. 
Other recent works~\cite{liang2025diffusion, he2026unirelight, yang2025unified, wang2024taming, feng2025dit4edit, long2025follow, shen2025follow, zhang2024lumisculpt, lin2026illumicraft, zeng2025lumen, xue2026georelight} further explore various types of lighting control within large-scale video generative models. However, all these methods operate in an offline, clip-level paradigm where the full lighting condition must be specified before generation, precluding real-time interactive adjustment. Our method instead performs streaming relighting, responding to dynamic light map inputs frame by frame.

\subsection{Real-time Video Generation}
Video diffusion models~\cite{blattmann2023stable, yang2024cogvideox, wan2025wan, chen2024diffusion, kong2024hunyuanvideo, huang2026self} produce high-quality and temporally coherent results, but their iterative denoising and full-clip computation make low-latency generation challenging. A line of work accelerates sampling by reducing the number of inference steps, including progressive distillation~\cite{salimans2022progressive}, consistency models~\cite{wang2023videolcm, zheng2024trajectory}, rectified flow~\cite{liu2022flow, esser2024scaling}, and shortcut models~\cite{frans2024one}, enabling few-step or even one-step generation. Beyond step reduction, another line of work designs streaming pipelines that produce frames continuously: StreamDiffusion~\cite{kodaira2025streamdiffusion} batches denoising for real-time image generation, StreamV2V~\cite{liang2025looking} extends this idea to video-to-video translation with a feature bank.

\subsection{Controllable Video Generation}
Controllable video generation~\cite{ma2024follow, zhang2024controlvideo,wan2025unipaint,ma2025followfaster, gao2026pai, song2024processpainter, song2026vista, song2026streamingeffect, xing2024make, he2024cameractrl, tian2024emo, zhao2024motiondirector, xing2025motioncanvas} synthesizes videos that follow user-specified conditions beyond text prompts, which is essential for creative workflows where text alone is too coarse for fine-grained user intent. To this end, a common approach is to inject auxiliary signals into pre-trained video diffusion models, with modalities ranging from structural maps and motion trajectories to identity and lighting~\cite{ma2025controllable}. Despite the variety of modalities, these methods share the same offline pipeline: the full conditioning sequence is fixed before generation, and the video is produced in a single pass. As a result, users cannot refine inputs based on intermediate results, which is impractical for iterative creative work. A more interactive paradigm appears in 3D-aware portrait relighting~\cite{zhang2021neural, cai2024real, mei2025lux}, where users can drag light directions and see immediate updates, yet these methods only handle portraits and rely on neural rendering. Our method brings such interactivity into diffusion-based streaming relighting: the user adjusts a 3D light source's position, intensity, and color in a reconstructed scene mesh, and the relit video is produced chunk-by-chunk following the live light state.

\section{Method}

\begin{figure*}[t]
    \centering
    \includegraphics[width=0.96\textwidth]{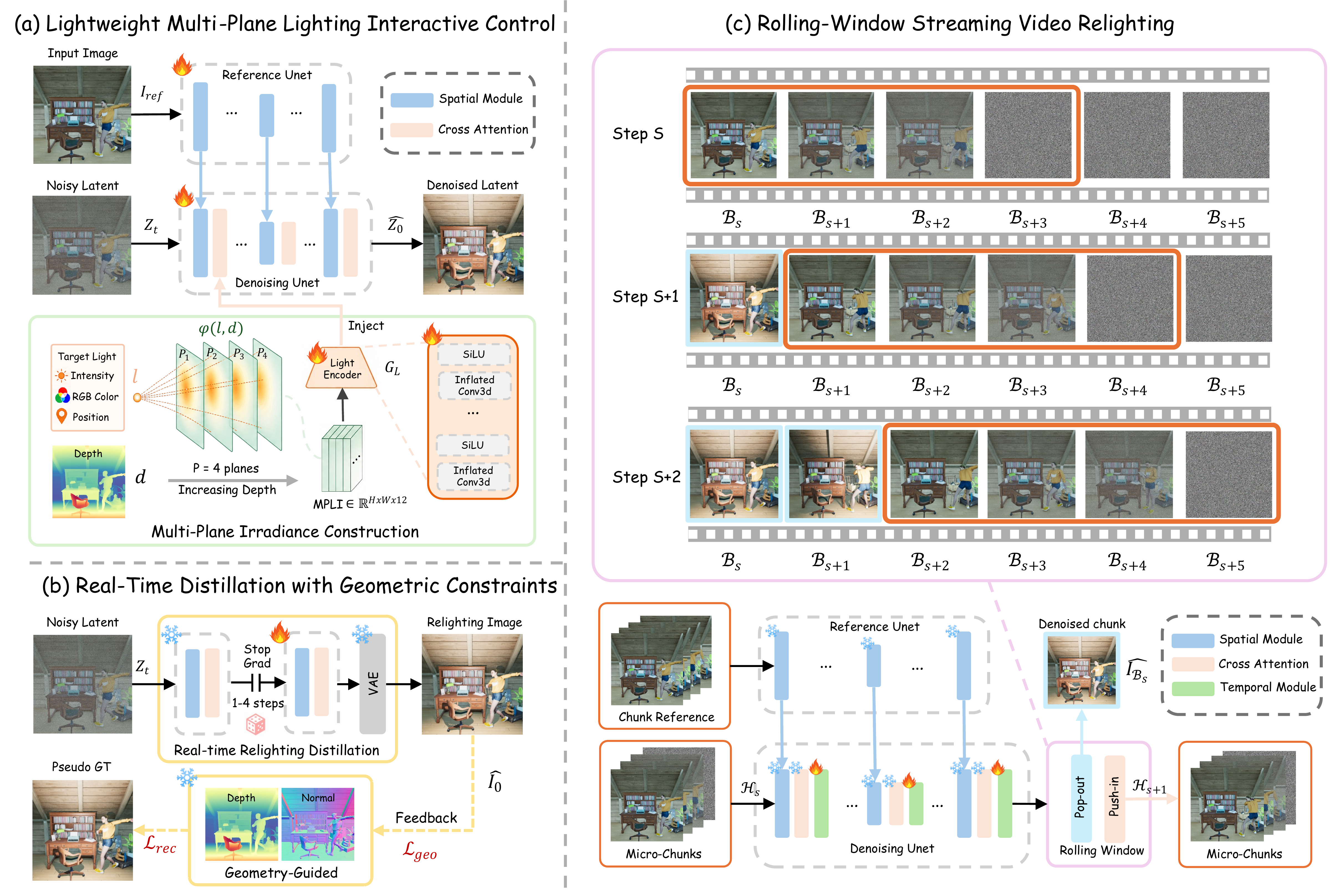}
    \caption{
    \textbf{Overview of LiveLight.}
    (a) We first perform lightweight multi-plane image relighting by constructing a multi-plane irradiance representation from the target light and estimated depth. The resulting MPLI condition is encoded by the light encoder and injected into the diffusion relighting backbone together with the reference image.
    (b) To improve structure-preserving relighting, we introduce a geometry-guided feedback stage, where intermediate relighting predictions are evaluated by a geometry estimator and supervised with reconstruction and geometry-constrained losses.
    (c) For video relighting, we train the temporal module using a rolling-window strategy. At each streaming step, the oldest denoised chunk is emitted, the remaining latent chunks are retained as temporal history, and a new noisy chunk is pushed into the rolling window. This enables temporal consistency while maintaining bounded latency.
    }
    \label{fig:framework}
\end{figure*}

We aim to generate geometrically plausible and temporally coherent relighting
results from a reference image or video. Given a reference
sequence \(I^{\mathrm{ref}}_{1:N}\) and a stream of target lighting
conditions \(\ell_{1:N}\), the objective is to synthesize a
relit sequence \(\hat{I}_{1:N}\) that follows the desired illumination
while preserving the subject identity, texture, and geometry. For image
relighting, \(N=1\); for video relighting, frames are generated in a
streaming and low-latency manner. Formally, each relit frame is generated by
combining reference appearance features and structured lighting features:
\begin{equation}
    \hat{I}_i
    =
    \mathcal{D}_{\theta}
    \left(
        Z_t
        \,\middle|\,
        R(I_i^{\mathrm{ref}}),
        G_L\!\left(\Phi(\ell_i, d_i)\right)
    \right),
    \quad i=1,\dots,N,
\end{equation}
where \(\mathcal{D}_{\theta}\) is the denoising backbone, \(Z_t\) denotes
the noisy latent at denoising step \(t\), \(R\) is the reference appearance
extractor, \(G_L\) is the light encoder, \(\Phi(\ell_i,d_i)\) denotes the
Multi-Plane Light Irradiance~\cite{bian2025relightmaster} condition constructed from the target lighting
\(\ell_i\) and the estimated subject depth \(d_i\), \(N\) denotes total frame number.

As shown in Fig.~\ref{fig:framework}, our method consists of three key
components. First, we introduce Lightweight Multi-Plane Lighting Control,
which provides an efficient and spatially aware representation for
interactive illumination editing (Sec.~\ref{sec:mpli}). Second, we perform
Real-Time Distillation with Geometric Constraints to accelerate inference
while improving geometry-consistent relighting
(Sec.~\ref{sec:physical_feedback}). Third, we develop Rolling-Window
Streaming Video Relighting to enable low-latency video generation with
per-frame reference refresh (Sec.~\ref{sec:streaming}).

\subsection{Relighting Data Synthesis}
\label{sec:data}

High-quality paired relighting data is difficult to capture in
real-world settings, as it requires controllable light sources, calibrated
cameras, and repeated recordings of the same subject under different
illumination conditions. We therefore construct a large-scale synthetic
relighting dataset using diverse 3D assets, motion and camera
trajectories, and physically based rendering. Our key design is to decouple
illumination from identity, pose, expression, camera motion, and scene
content, enabling the model to learn lighting-dependent appearance changes
rather than fixed subject-light correlations.
\begin{figure}[t]
    \centering
    \includegraphics[width=\linewidth]{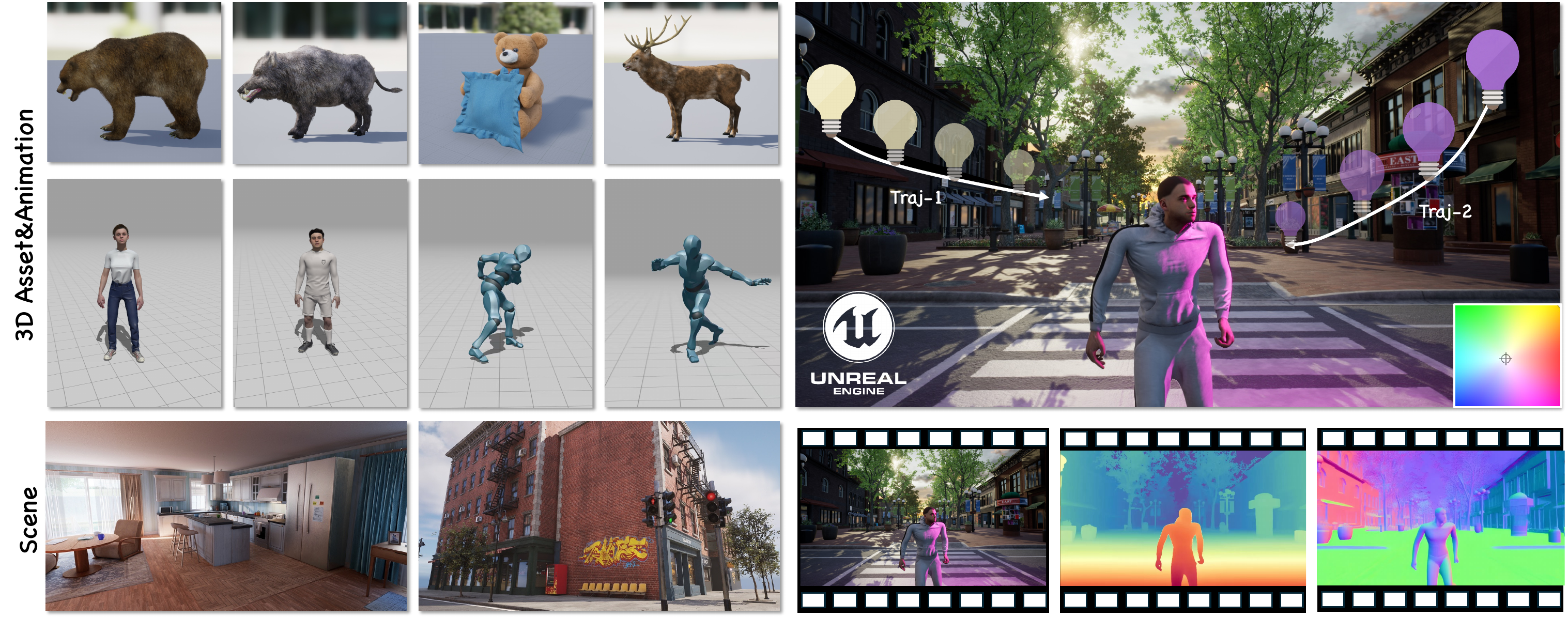}
    \caption{
   \textbf{Overview of our synthetic relighting data synthesis pipeline.}
    We randomly combine diverse 3D assets (humans, animals, and objects) with animations and indoor/outdoor scenes, and render them in Unreal Engine~5 under controllable point lights with varying positions, intensities, and colors. For each frame, the renderer exports the RGB image along with depth map and surface normal map for geometry-aware supervision.
    }
    \label{fig:data_synthesis}
\end{figure}

As shown in Fig.~\ref{fig:data_synthesis}, we randomly combine 120 diverse
3D assets, 33 motion/camera trajectories, and 20 scenes to render
paired relighting videos. For each sampled motion sequence, we render a
canonical reference video under neutral illumination and 10 target videos
under different controllable lights. The reference and target videos share
the same identity, pose, expression, camera path, and scene content, and
differ only in illumination. Each frame is exported with a 
depth map and a surface normal map, which are used for MPLI construction and
geometry-aware supervision. After automatic visual-quality screening and
manual verification, the final dataset contains approximately 250K relighting
video sequences. Target lighting conditions are sampled from a controllable parametric space, including image-space position, relative depth, intensity, and RGB color. We
cover both colored and white illumination with an approximate ratio of 6:4.
The light intensity is randomly sampled within \([0.5, 1.5]\) in normalized
UE units. To avoid degenerate near-black colored lights, we sample colored
illumination from a high-saturation and high-value HSV space before converting
it to RGB. For image-level relighting, individual reference-target frame pairs
are sampled from these sequences. For video relighting, we generate temporally
smooth lighting trajectories so that the reference and target videos provide
consistent supervision over time. We further apply domain randomization over
backgrounds, ambient illumination, material, camera exposure, and
color temperature to improve robustness to real-world videos.


\subsection{Multi-Plane Lighting Interactive Control}
\label{sec:mpli}

A central challenge in video relighting is how to represent the target
illumination in a compact, controllable, and spatially meaningful form.
Instead of relying on a low-dimensional lighting code or an expensive dense
illumination map, we adopt a Multi-Plane Light Irradiance (MPLI)~\citep{bian2025relightmaster} condition to
provide explicit control over light position, relative depth, intensity,
and color. Our contribution is a lightweight lighting-control branch that
encodes this structured condition and injects it into the relighting
backbone, enabling interactive illumination with low computational
overhead.

\noindent{MPLI-based lighting condition.}
Given the target light \(\ell_i\) and the estimated subject depth \(d_i\),
we construct an MPLI condition by projecting the user-specified light onto
a small set of fronto-parallel depth planes around the subject:
\begin{equation}
    \Phi(\ell_i, d_i)
    =
    \operatorname{Concat}_{p=1}^{P}
    \phi_p(\ell_i, d_i)
    \quad
    \in \mathbb{R}^{H \times W \times 3P}
\end{equation}
where \(\phi_p(\ell_i,d_i)\) denotes the RGB irradiance map on the
\(p\)-th plane. We use \(P=4\), producing a 12-channel lighting condition.
This compact tensor preserves spatially varying illumination cues while
allowing users to directly edit the light position, intensity, and
color.

\noindent{Lightweight light adapter.}
Although the MPLI condition is compact, it still contains spatially aligned
lighting information that must be adapted to the feature space of the
relighting backbone. We therefore design a lightweight light encoder
\(G_L\), implemented with a small stack of convolutional blocks, to map the
MPLI tensor into diffusion-compatible lighting features. The encoded light
features are fused with the reference appearance features by injection:
\begin{equation}
    \widetilde{E}_i
    =
    E_\theta
    \left(
        R(I_i^{\mathrm{ref}})
    \right)
    +
    G_L
    \left(
        \Phi(\ell_i, d_i)
    \right),
\end{equation}
where \(E_\theta(\cdot)\) denotes the backbone encoder feature,
\(R(\cdot)\) extracts appearance feature, and
\(\widetilde{E}_i\) is the resulting light-aware feature. Specifically, \(G_L\) is realized as a lightweight adapter with only
\(\sim\!1\)M parameters. It transforms the MPLI condition into compact
lighting tokens, which interact with the diffusion features through
cross-attention in shallow layers. This localized injection strategy
introduces spatially varying illumination cues at an early stage, while
avoiding unnecessary perturbation to deeper denoising blocks that encode
appearance priors, thereby preserving identity and texture fidelity.

This design separates appearance preservation from lighting control: the
reference branch provides identity and texture cues, while the lightweight
MPLI branch provides spatial illumination guidance. Since the lighting
condition is explicitly parameterized by user-controllable attributes,
changing the target light only requires updating \(\Phi(\ell_i, d_i)\) and passing it
through the same lightweight encoder, which enables interactive
control over diverse lighting.

\begin{figure}[t]
    \centering
    \includegraphics[width=\linewidth]{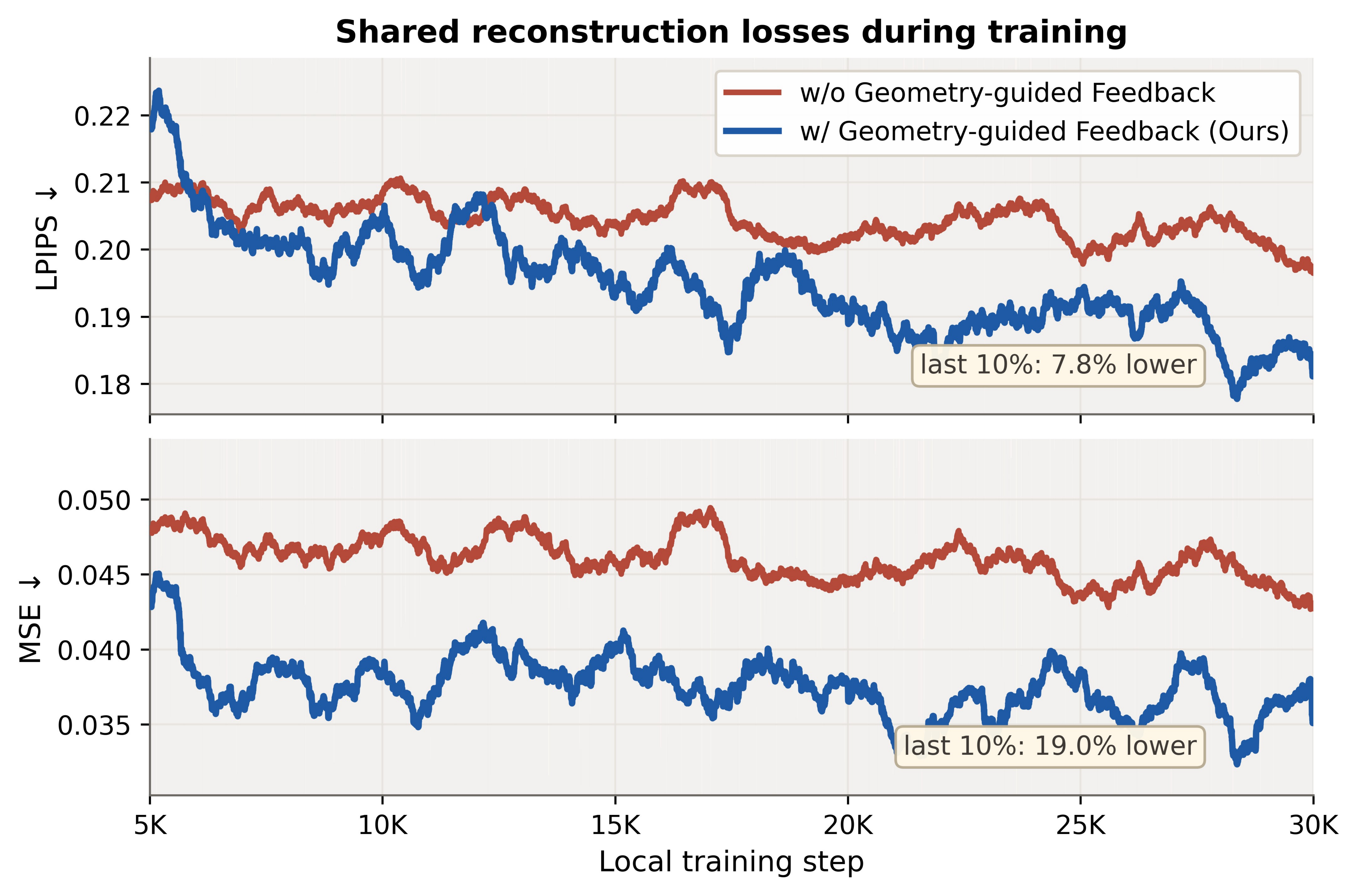}
    \caption{
    \textbf{Geometry-guided feedback for structure-preserving relighting.}
    Compared with the baseline without geometry-guided feedback, our model maintains lower LPIPS and MSE, especially in the later training stage. Over the last \(10\%\) of steps, these metrics are \(7.8\%\) and \(19.0\%\) better, respectively. This indicates that depth-normal feedback provides an effective geometric regularization toward consistent relighting.
    }
    \label{fig:physical_feedback}
\end{figure}

\subsection{Real-Time Distillation with Geometric Constraints}
\label{sec:physical_feedback}

The lightweight lighting branch provides efficient and controllable
illumination guidance, but standard diffusion sampling still requires
multiple denoising steps, which limits real-time interaction. We observe
that video relighting contains substantial denoising redundancy: once
conditioned on the reference appearance and MPLI lighting, the earliest
denoising steps already determine the main light direction and global
exposure, place shadows or highlights on the face, and establish
depth-dependent falloff, while the remaining iterations mostly refine
texture and high-frequency details. This motivates us to replace the
redundant multi-step refinement process with a compact few-step sampling
schedule.

\noindent{Few-step relighting distillation.}
To make the compact schedule effective at inference, we train the relighting
model directly with the same four-step rollout used at test time. Starting
from noised latent, the denoising backbone follows a fixed few-step
timestep schedule. For each training sample, we randomly choose a terminal update
\(m \in \{1,2,3,4\}\), detach the preceding \(m-1\) updates that construct
the intermediate latent, and back-propagate only through the final update.
The predicted clean latent is decoded by the frozen VAE and supervised
against the target relit image using reconstruction and geometry losses,
which exposes the model to different positions in the compact trajectory
while keeping memory cost low.

\noindent{Geometry-guided supervision.}
Relighting should change illumination while preserving geometry.
As shown in Fig.~\ref{fig:physical_feedback}, incorporating depth and normal
feedback regularizes the few-step model toward more stable, geometry-preserving
relighting. Given the generated image \(\hat{I}_i\), a frozen geometry estimator
\(\Psi\) predicts depth and normal cues, which are compared with
precomputed ground-truth geometry:
\begin{equation}
    \begin{aligned}
    \mathcal{L}_{\mathrm{geo}}
    =
    &\left\|
        \Psi_D(\hat{I}_i) - d_i^{\ast}
    \right\|_1
    +
    \lambda_N
    \left\|
        1 -
        \left\langle
            \Psi_N(\hat{I}_i), n_i^{\ast}
        \right\rangle
    \right\|_1 .
    \end{aligned}
\end{equation}
Here  \(d_i^{\ast}\) and \(n_i^{\ast}\) are
pseudo ground-truth depth and normal, and \(\Psi_D,\Psi_N\) are predictions
from the frozen estimator. Gradients are preserved with respect to
\(\hat{I}_i\), allowing geometry errors to update the relighting model.
This branch is used only during training and adds no inference cost.

\subsection{Rolling-Window Streaming Video Relighting}
\label{sec:streaming}

Frame-wise relighting can achieve low latency, but it lacks temporal context
and often produces flickering artifacts. In contrast, full-clip video
diffusion provides stronger temporal consistency but requires the entire
sequence before generation, making it unsuitable for streaming applications.
We therefore train the temporal module with a rolling-window
strategy. The model only observes a short window of micro-chunks at each
step, while previously processed latent chunks are propagated as the
streaming history \(\mathcal{H}\).

Let \(\mathcal{B}_s\) denote the \(s\)-th micro-chunk and let \(K\) be the
number of micro-chunks maintained in the rolling window. At streaming step
\(s\), the active history consists of \(K\) latent chunks. The video denoising
backbone emits the oldest chunk, keeps the remaining chunks as temporal
context, and appends a new noisy chunk to the tail:
\begin{equation}
    \left(
        \hat{I}_{\mathcal{B}_s},
        \mathcal{H}_{s+1}
    \right)
    =
    \mathcal{D}_{\theta}
    \left(
        \mathcal{H}_{s}
        \,\middle|\,
        \left\{
            R(I_i^{\mathrm{ref}}),
            G_L\!\left(\Phi(\ell_i, d_i)\right)
        \right\}_{i \in \mathcal{W}_s}
    \right),
    \label{eq:rolling_window}
\end{equation}

Here \(\mathcal{H}_{s}\) is the chunk-level streaming history for window \(\mathcal{W}_s\).
After each step, the oldest micro-chunk is decoded and emitted, the window
slides forward by one chunk, and a new noisy chunk is inserted for future
generation. Since each chunk is generated only once, the method avoids
overlap-based re-generation and blending, while still allowing temporal
information to flow through the retained latent context. Details are provided in
Appendix~\ref{app:rolling_stream}.

The window length \(K\) controls the latency-consistency trade-off. A larger
window provides longer temporal context and improves consistency, but
increases memory usage and output latency. A smaller window reduces latency
but weakens temporal modeling, which may lead to flickering or inconsistent
lighting changes. We therefore use a moderate window size during video
adaptation and keep the same rolling-window schedule for inference, reducing
the train-test mismatch for streaming generation.

\subsection{Training Objectives}
\label{sec:training_objectives}

We train the proposed framework in three stages, corresponding to the
modules introduced in Sec.~\ref{sec:mpli}, Sec.~\ref{sec:physical_feedback},
and Sec.~\ref{sec:streaming}.

\noindent{Stage 1: Image-level relighting.}
We first train the ReferenceNet-based image relighting model with the MPLI
condition introduced in Sec.~\ref{sec:mpli}. Given a paired reference and
target relit image, the target image is encoded into latent space,
perturbed with Gaussian noise, and reconstructed by the denoising network
conditioned on reference appearance and MPLI lighting. The objective is
\begin{equation}
    \mathcal{L}_{\mathrm{stage1}}
    =
    \mathbb{E}_{\tau,\epsilon}
    \left[
        \|\hat{z}^0-z^0\|_2^2
    \right].
\end{equation}

\noindent{Stage 2: Geometry-plausible refinement.}
In the second stage, we enable the geometry-guided feedback branch and refine the
relighting model with geometry-aware supervision. The training objective is
\begin{equation}
    \mathcal{L}_{\mathrm{stage2}}
    =
    \mathcal{L}_{\mathrm{rec}}
    +
    \lambda_{\mathrm{geo}}
    \mathcal{L}_{\mathrm{geo}}.
\end{equation}
Here \(\mathcal{L}_{\mathrm{geo}}\) denotes the geometry-aware
feedback loss defined in Sec.~\ref{sec:physical_feedback}. This stage improves
geometry-consistent shading without adding any inference-time overhead.

\noindent{Stage 3: Streaming video adaptation.}
Finally, we adapt the image model to streaming video relighting using the
rolling-window strategy in Sec.~\ref{sec:streaming}. During training, we
simulate the same rolling-window process as inference. Intermediate
windows are advanced without gradient computation, and the emitted frames
are supervised using RGB and perceptual losses:
\begin{equation}
    \mathcal{L}_{\mathrm{stage3}}
    =
    \frac{1}{|\mathcal{S}_s|}
    \sum_{i\in\mathcal{S}_s}
    \left[
        \lambda_{\mathrm{mse}}
        \|\hat{I}_i-I_i^{\mathrm{gt}}\|_2^2
        +
        \lambda_{\mathrm{lpips}}
        \mathcal{L}_{\mathrm{lpips}}
        (\hat{I}_i,I_i^{\mathrm{gt}})
    \right],
\end{equation}
where \(\mathcal{S}_s\) denotes the supervised emitted frames.

\begin{figure*}[t]
  \centering
\includegraphics[width=\linewidth]{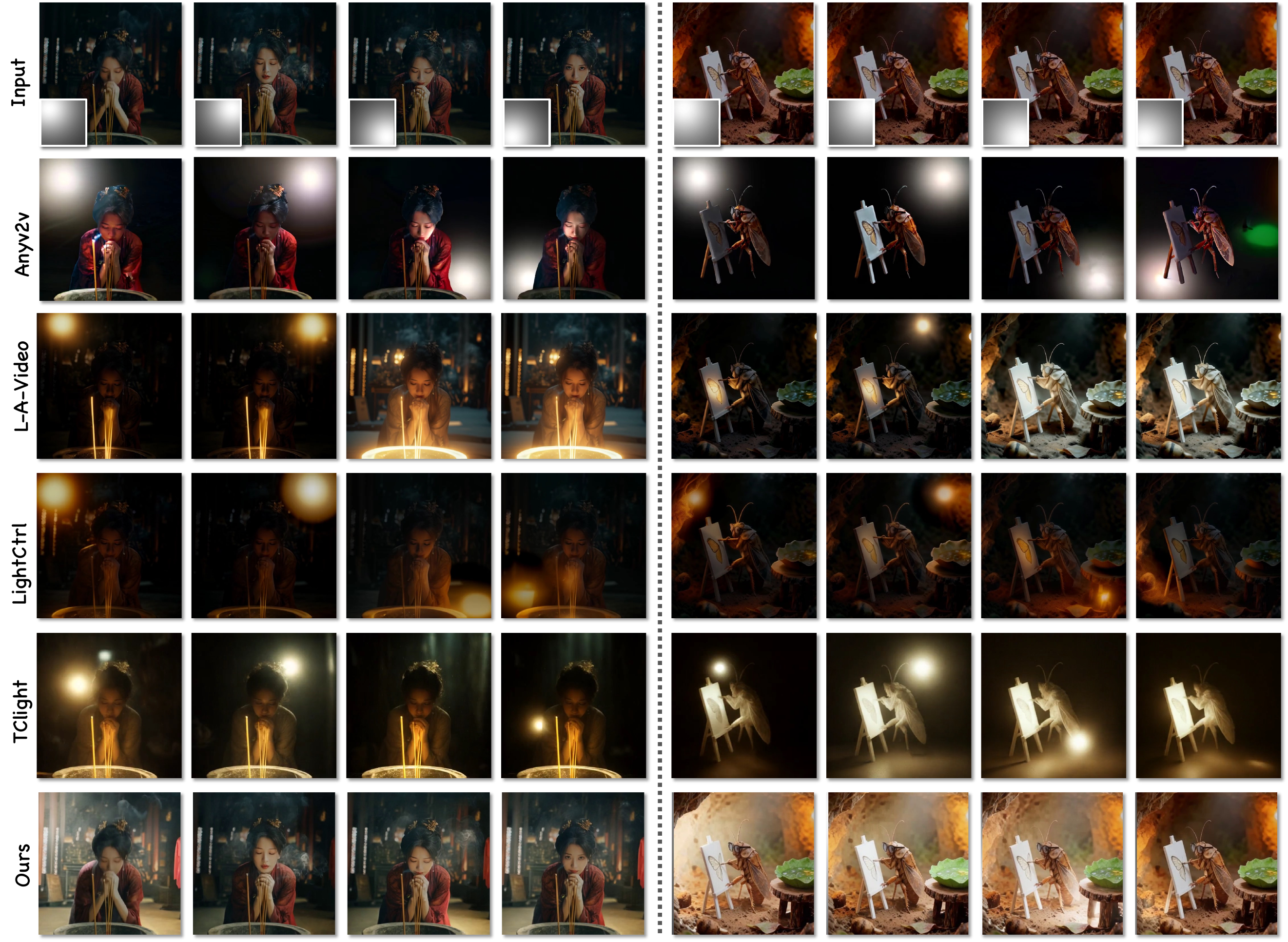}
  \caption{\textbf{Comparison of previous lighting control methods.} LiveLight outperforms existing methods in terms of controllability, temporal stability, and real-time efficiency.}
  \label{fig:comparison}
\end{figure*}

\section{Experiments}

\subsection{Application}

\begin{figure*}[t]
  \centering
  \includegraphics[width=\linewidth]{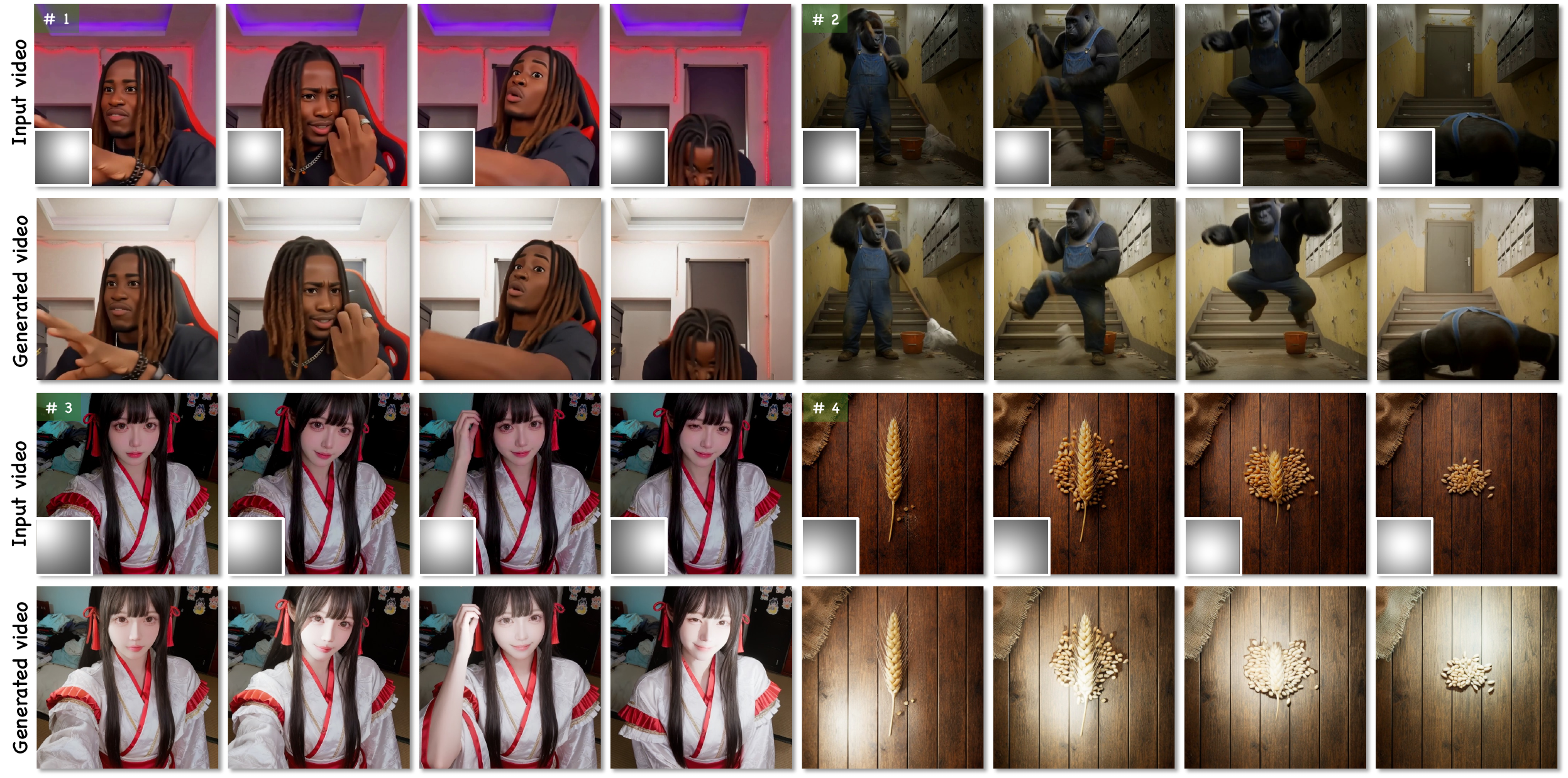}
  \caption{\textbf{Visual results of real-time lighting control with LiveLight.}
  LiveLight enables high-quality relighting video generation under target
  lighting while preserving the content and motion of the input video.}
  \label{fig:gallery1}  
\end{figure*}

\noindent \textbf{Real-time lighting control.}
LiveLight enables real-time generation of controllable relighting videos from
ordinary input videos. Benefiting from the proposed rolling-window
strategy and per-frame reference refresh, our method can continuously update
the lighting condition with low latency while maintaining temporal coherence.
As shown in Fig.~\ref{fig:gallery1}, LiveLight follows the target lighting
and preserves scene content, motion, and visual quality across diverse videos.

\begin{figure}[t]
  \centering
  \includegraphics[width=0.95\linewidth]{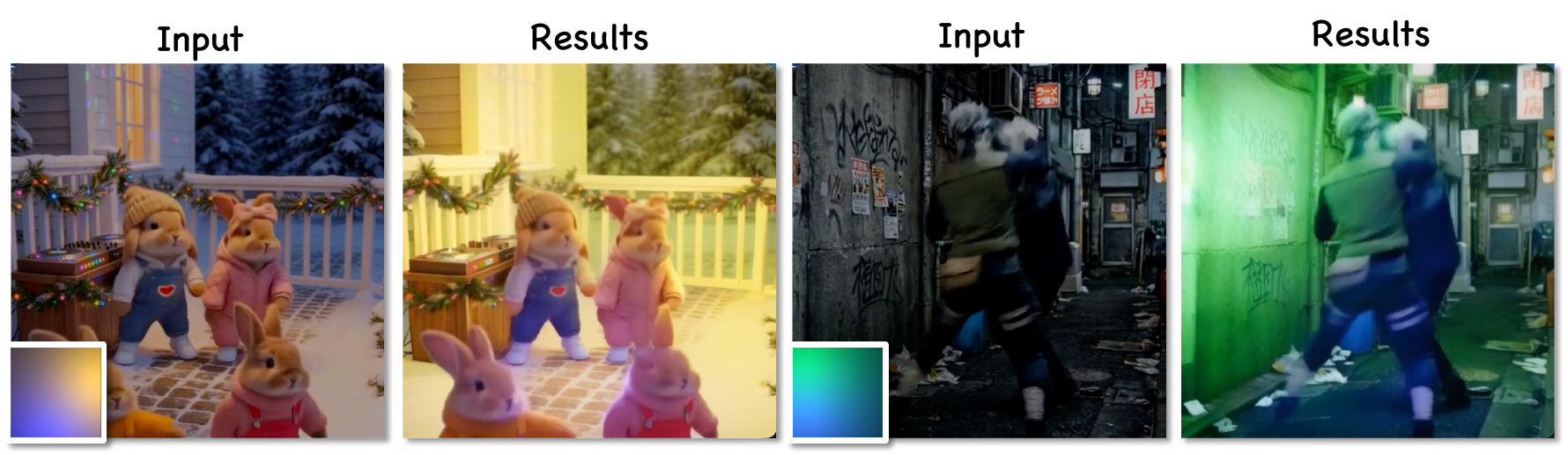}
  \caption{\textbf{Visual results of multiple color control.} LiveLight supports simultaneous control of multiple colored light sources, producing spatially distinct and geometrically plausible relighting effects.}
  \label{fig:multicolor}
\end{figure}

\noindent \textbf{Diverse color control.}
LiveLight also supports real-time control over diverse lighting colors and
positions. The proposed multi-plane light representation provides spatially
meaningful lighting guidance, allowing users to specify colored lighting
with different locations and intensities. Fig.~\ref{fig:gallery2} shows that
our method can produce vivid colored lighting effects while maintaining
coherent video appearance across frames. Furthermore, as shown in Fig.~\ref{fig:multicolor}, LiveLight can simultaneously handle multiple colored light sources, producing spatially distinct relighting effects within a single frame.

\begin{figure}[t]
  \centering
  \includegraphics[width=\linewidth]{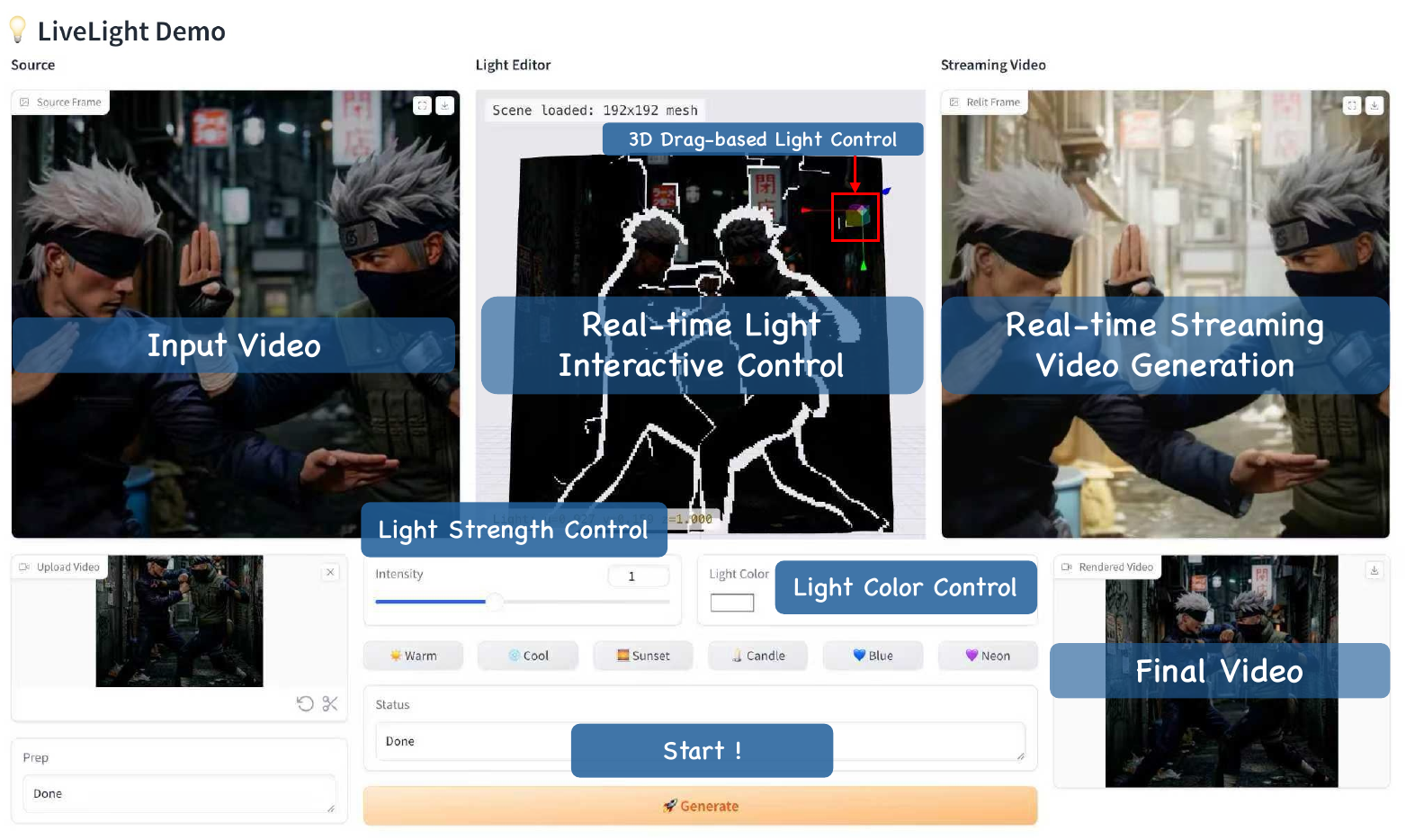}
  \caption{\textbf{Illustration of LiveLight's user interface.}  The interface allows users to interactively specify lighting conditions and visualize the relighting results in real time. Users can adjust light color, intensity, and position through intuitive controls, enabling flexible and dynamic lighting effects on the input video.}
  \label{fig:gui}  
\end{figure}

\noindent \textbf{User Interface.} We design a user interface for LiveLight to allow users to interactively specify lighting conditions and visualize the relighting results in real time. As shown in Fig.~\ref{fig:gui}, the interface provides intuitive controls for adjusting light color, intensity, and position, enabling users to create flexible and dynamic lighting effects on their videos. The real-time feedback allows users to see the impact of their adjustments immediately, enhancing the creative process and making it easier to achieve the desired relighting results.

\subsection{Implementation Details}

In our implementation, we adopt a three-stage training pipeline to optimize the components of LiveLight. In Stage~1, we conduct image-level relighting training. Both the denoising U-Net and the ReferenceNet are initialized from pretrained Stable Diffusion weights~\citep{rombach2022high}. This stage is trained for 50K iterations with a batch size of 32, and all model parameters are updated. In Stage~2, we enable the geometry-guided feedback branch and fine-tune the relighting model with geometry-aware supervision. This stage is trained for 20K iterations with a batch size of 16, keeping the geometry estimator frozen. In Stage~3, we insert temporal attention layers initialized from AnimateDiff~\citep{guo2023animatediff} motion module weights and train the streaming video adaptation with the rolling-window strategy. The chunk size is set to $M=4$. This stage is trained for 30K iterations with a batch size of 8. All training is conducted on 8 NVIDIA H20 GPUs.
We use AdamW with weight decay $0.01$. The learning rate is $1\times10^{-5}$ for Stages~1 and~2, and $5\times10^{-6}$ for Stage~3. Our training data is a self-constructed large-scale dataset of UE5-rendered video sequences with diverse lighting conditions, covering various identities, poses, and lighting trajectories. More details can be found in the supplementary materials.

\begin{table*}[t]
  \centering
  \caption{\textbf{Quantitative Comparisons with Other Methods.}
  \textcolor{Red}{\textbf{Red}} and \textcolor{Blue}{\textbf{Blue}} denote the best and second best results, respectively.}
  \label{tab:quantitative_comparison}
  \resizebox{\linewidth}{!}{%
    \begin{tabular}{l|ccccc|cccc}
      \toprule
      \multirow{2}{*}{Method}
        & \multicolumn{5}{c|}{Quality Metrics}
        & \multicolumn{4}{c}{VBench Metrics~\citep{huang2024vbench}} \\
      \cmidrule(lr){2-6}\cmidrule(lr){7-10}
        & Aesthetic $\uparrow$
        & FVD $\downarrow$
        & PSNR$_{y}$ $\uparrow$
        & PSNR$_{light}$ $\uparrow$
        & Time (s) $\downarrow$
        & Sub. Cons. $\uparrow$
        & Back. Cons. $\uparrow$
        & Motion Smooth. $\uparrow$
        & Temporal Cons. $\uparrow$ \\
      \midrule
      AnyV2V~\citep{zhang2025scaling}
        & 0.4929 & 1082.6 & 11.063 & 15.923 & 89
        & 0.7980 & 0.8970 & 0.9707 & 0.9656 \\
      LightCtrl~\citep{peng2026lightctrl}
        & 0.5307 & 987.3 & 11.782 & 18.486 & 122
        & 0.8696 & 0.9070 & 0.9728 & 0.9559 \\
      Light-A-Video~\citep{zhou2025light}
        & \textcolor{Blue}{\textbf{0.5533}} & 1053.8 & 11.247 & 17.612 & 115
        & 0.8678 & 0.9244 & \textcolor{Blue}{\textbf{0.9782}} & 0.9740 \\
      TC-Light~\citep{liu2026tc}
        & 0.5170 & \textcolor{Blue}{\textbf{862.4}} & \textcolor{Blue}{\textbf{12.136}} & \textcolor{Blue}{\textbf{19.274}} & \textcolor{Blue}{\textbf{38}}
        & \textcolor{Blue}{\textbf{0.9402}} & \textcolor{Blue}{\textbf{0.9502}} & 0.9642 & \textcolor{Blue}{\textbf{0.9799}} \\
      \midrule
      LiveLight (Ours)
        & \textcolor{Red}{\textbf{0.5748}} & \textcolor{Red}{\textbf{738.5}} & \textcolor{Red}{\textbf{12.847}} & \textcolor{Red}{\textbf{21.538}} & \textcolor{Red}{\textbf{1.01}}
        & \textcolor{Red}{\textbf{0.9561}} & \textcolor{Red}{\textbf{0.9687}} & \textcolor{Red}{\textbf{0.9856}} & \textcolor{Red}{\textbf{0.9872}} \\
      \bottomrule
    \end{tabular}%
  }
\end{table*}

\begin{table}[t]
  \centering
  \caption{\textbf{Quantitative Comparisons with Synthetic data.}
  \textcolor{Red}{\textbf{Red}} and \textcolor{Blue}{\textbf{Blue}} denote the best and second best results, respectively.}
  \label{tab:synthetic_comparison}
  \resizebox{0.8\linewidth}{!}{%
    \begin{tabular}{l|ccc}
      \toprule
      \multirow{2}{*}{Method}
        & \multicolumn{3}{c}{Synthetic Data Metrics} \\
      \cmidrule(lr){2-4}
        & PSNR $\uparrow$
        & SSIM $\uparrow$
        & LPIPS $\downarrow$ \\
      \midrule
      AnyV2V~\citep{zhang2025scaling}
        & 31.482 & 0.8936 & 0.2693 \\
      LightCtrl~\citep{peng2026lightctrl}
        & 35.617 & 0.9254 & 0.2148 \\
      Light-A-Video~\citep{zhou2025light}
        & 37.385 & 0.9418 & 0.1726 \\
      TC-Light~\citep{liu2026tc}
        & \textcolor{Blue}{\textbf{39.726}} & \textcolor{Blue}{\textbf{0.9571}} & \textcolor{Blue}{\textbf{0.1387}} \\
      \midrule
      LiveLight (Ours)
        & \textcolor{Red}{\textbf{43.264}} & \textcolor{Red}{\textbf{0.9778}} & \textcolor{Red}{\textbf{0.1064}} \\
      \bottomrule
    \end{tabular}%
  }
\end{table}

\begin{table*}[t]
  \centering
  \caption{\textbf{Quantitative comparison on longer videos with VBench~\citep{huang2024vbench}.}
  $\uparrow$ indicates that higher is better.
  \textcolor{Red}{\textbf{Red}} and \textcolor{Blue}{\textbf{Blue}} denote the best and second best results, respectively.}
  \label{tab:long_video_comparison}
  \setlength\tabcolsep{3pt}
  \resizebox{\linewidth}{!}{%
    \begin{tabular}{l|cccc|cccc}
      \toprule
      \multirow{2}{*}{Method}
        & \multicolumn{4}{c|}{256 frames}
        & \multicolumn{4}{c}{512 frames} \\
      \cmidrule(lr){2-5}\cmidrule(lr){6-9}
        & Sub. Cons. $\uparrow$
        & Back. Cons. $\uparrow$
        & Aes. Qual. $\uparrow$
        & Motion Smooth. $\uparrow$
        & Sub. Cons. $\uparrow$
        & Back. Cons. $\uparrow$
        & Aes. Qual. $\uparrow$
        & Motion Smooth. $\uparrow$ \\
      \midrule
      AnyV2V~\citep{zhang2025scaling}
        & 0.7728 & 0.8706 & 0.4671 & 0.9534
        & 0.7548 & 0.8546 & 0.4481 & 0.9364 \\
      LightCtrl~\citep{peng2026lightctrl}
        & 0.8412 & 0.8823 & 0.5057 & 0.9563
        & 0.8262 & 0.8633 & 0.4887 & 0.9403 \\
      Light-A-Video~\citep{zhou2025light}
        & 0.8386 & 0.8978 & \textcolor{Blue}{\textbf{0.5263}} & \textcolor{Blue}{\textbf{0.9617}}
        & 0.8216 & 0.8828 & \textcolor{Blue}{\textbf{0.5083}} & \textcolor{Blue}{\textbf{0.9477}} \\
      TC-Light~\citep{liu2026tc}
        & \textcolor{Blue}{\textbf{0.9128}} & \textcolor{Blue}{\textbf{0.9247}} & 0.4918 & 0.9472
        & \textcolor{Blue}{\textbf{0.8972}} & \textcolor{Blue}{\textbf{0.9067}} & 0.4758 & 0.9283 \\
      \midrule
      LiveLight (Ours)
        & \textcolor{Red}{\textbf{0.9456}} & \textcolor{Red}{\textbf{0.9542}} & \textcolor{Red}{\textbf{0.5592}} & \textcolor{Red}{\textbf{0.9812}}
        & \textcolor{Red}{\textbf{0.9396}} & \textcolor{Red}{\textbf{0.9492}} & \textcolor{Red}{\textbf{0.5552}} & \textcolor{Red}{\textbf{0.9782}} \\
      \bottomrule
    \end{tabular}%
  }
\end{table*}

\subsection{Comparison with baselines}

\begin{figure*}[t]
  \centering
  \includegraphics[width=\linewidth]{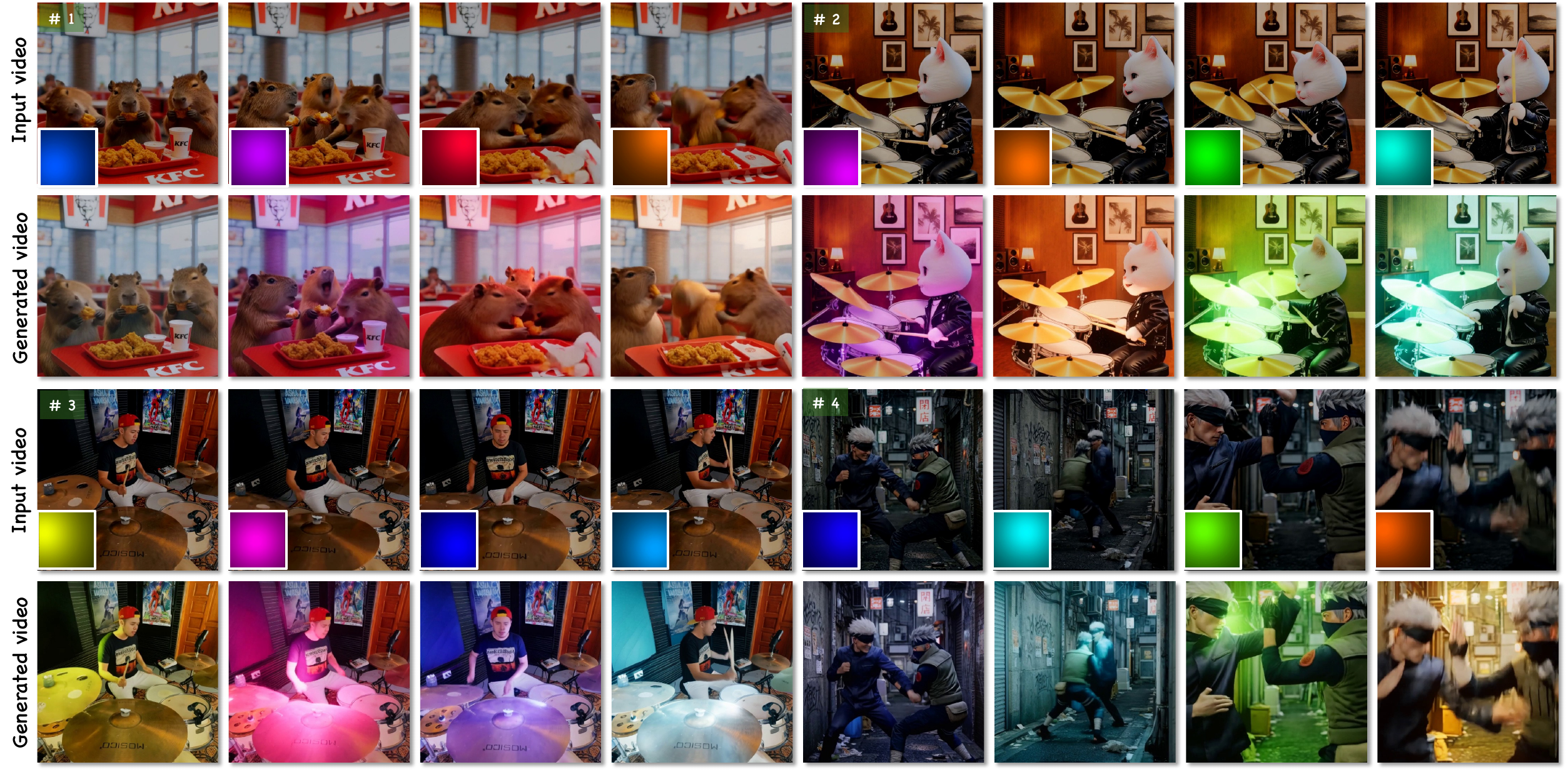}
  \caption{\textbf{Visual results of diverse color lighting control with
  LiveLight.} LiveLight supports real-time lighting control with different
  colors and positions, producing temporally coherent relighting results across
  dynamic videos.}
  \label{fig:gallery2}
\end{figure*}

\noindent \textbf{Qualitative comparison.}
Fig.~\ref{fig:gallery1} and Fig.~\ref{fig:gallery2} present qualitative results
of LiveLight on diverse Internet videos. Compared with image-based relighting
pipelines, LiveLight avoids obvious frame-wise flickering and produces more
stable lighting changes over time. Compared with offline video relighting
methods, LiveLight supports frame-by-frame control of light position, color, and
intensity in a streaming manner. Benefiting from the proposed MPLI
representation and rolling-window strategy, LiveLight follows the target
lighting condition while preserving subject appearance, scene structure, and
motion details. More visual comparisons are provided in the supplementary
material. 

\begin{figure*}[t]
  \centering
  \includegraphics[width=0.95\linewidth]{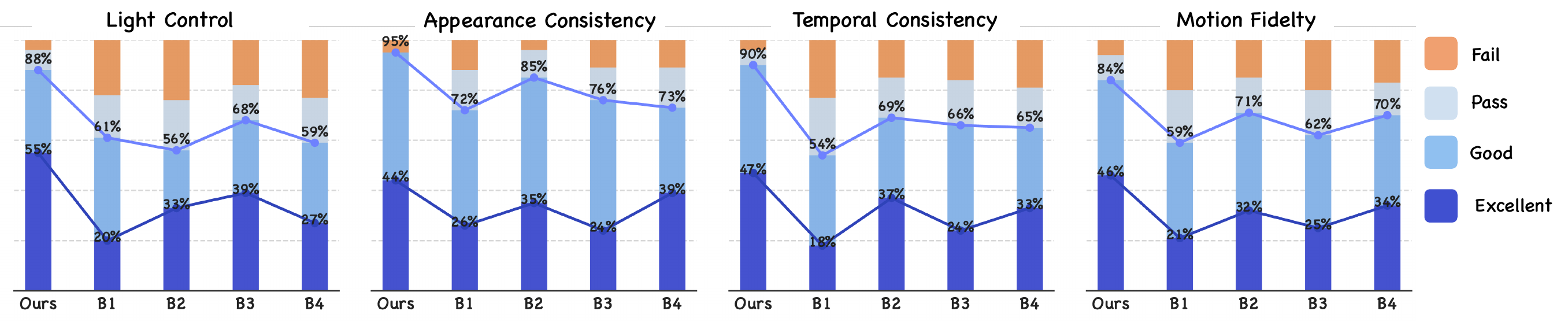}
  \caption{\textbf{User study.} We conduct a user study to evaluate the perceptual quality of relighting videos generated by LiveLight and baseline methods. Participants are asked to rate each result in terms of light control, appearance consistency, temporal consistency, and motion fidelity. LiveLight receives the highest preference across all criteria, demonstrating its ability to produce controllable, visually consistent, and temporally coherent relighting videos.}
  \label{fig:user_study}
\end{figure*}

\noindent \textbf{Quantitative comparison.}
We compare LiveLight with representative image-based and video-based relighting
baselines, including AnyV2V~\citep{zhang2025scaling},
LightCtrl~\citep{peng2026lightctrl},
Light-A-Video~\citep{zhou2025light}, and TC-Light~\citep{liu2026tc}. Following
the standard setting of previous video relighting methods, we collect 200
lighting-control videos from the Internet, and evaluate all methods on the same
16-frame clips.  For AnyV2V~\citep{zhang2025scaling}, we edit the first frame using IC-Light~\citep{zhang2025scaling} and feed it into the model.  Following prior video relighting work~\citep{peng2026lightctrl}, we first use five
task-level metrics. (1) \textbf{\textit{Aesthetic}}: we use an aesthetic predictor to
measure the overall visual appeal of generated videos. (2) \textbf{\textit{FVD}}: we
compute the distribution distance between generated videos and input videos to
evaluate video realism. (3) \textbf{\textit{PSNR$_y$}}: we compute PSNR on the luminance
channel within the light-map mask region against a pure-white reference video,
which measures whether the specified region receives stronger illumination.
(4) \textbf{\textit{PSNR$_{light}$}}: we directly overlay the target light map on the input
video and compute PSNR between this overlay reference and the generated result,
which evaluates whether the relighting follows the light trajectory while
preserving the foreground appearance. (5) \textbf{\textit{Time}}: we report the inference
cost of each method. 

We further adopt four VBench metrics~\citep{huang2024vbench} for a broader
assessment of video quality. \textbf{\textit{Subject Consistency}} evaluates whether the
foreground subject is preserved across frames. \textbf{\textit{Background Consistency}}
measures the coherence of static scene regions. \textbf{\textit{Motion Smoothness}}
evaluates inter-frame motion continuity, and \textbf{\textit{Temporal Consistency}}
measures overall frame-to-frame coherence. As shown in
Table~\ref{tab:quantitative_comparison}, LiveLight achieves the best overall
performance across both relighting-specific metrics and general video-quality
metrics, demonstrating superior controllability, temporal stability, and
real-time efficiency.

\noindent{Longer video evaluation.}
We further evaluate long-video relighting on another 200 Internet videos. All
methods are tested under two longer settings, 256 frames and 512 frames. We use the same four VBench metrics as in the main quantitative comparison. As reported in
Table~\ref{tab:long_video_comparison}, LiveLight consistently outperforms
competing methods under both video lengths, showing that our streaming design
better preserves stable relighting over extended temporal horizons.

\noindent{Synthetic data evaluation.} 
To evaluate the relighting quality under controlled conditions, we also test all methods on synthetic videos with known ground-truth lighting changes. We use PSNR, SSIM, and LPIPS to measure the fidelity of relighting results against the ground truth. We synthesized 100 videos in UE5 with various lighting trajectories and 10 different scene to comprehensively assess the performance of each method. Notably, all evaluation scenes and lighting conditions are unseen during training. As shown in Table~\ref{tab:synthetic_comparison}, LiveLight achieves the best performance across all metrics, demonstrating its ability to accurately follow target lighting conditions while preserving visual quality.

\noindent{User study.}
To account for the limitations of automatic metrics in capturing real-world
preferences, we conduct a user study with 20 volunteers on 50 randomly selected
test cases. The volunteers categorize each result into one of four levels:
excellent, good, pass, or fail, based on lighting accuracy, appearance
consistency, temporal coherence, and motion fidelity. The results are presented
in Fig.~\ref{fig:user_study}. LiveLight outperforms competing methods in both
automated metrics and human subjective preferences.

\begin{figure}[t]
  \centering
  \includegraphics[width=0.95\linewidth]{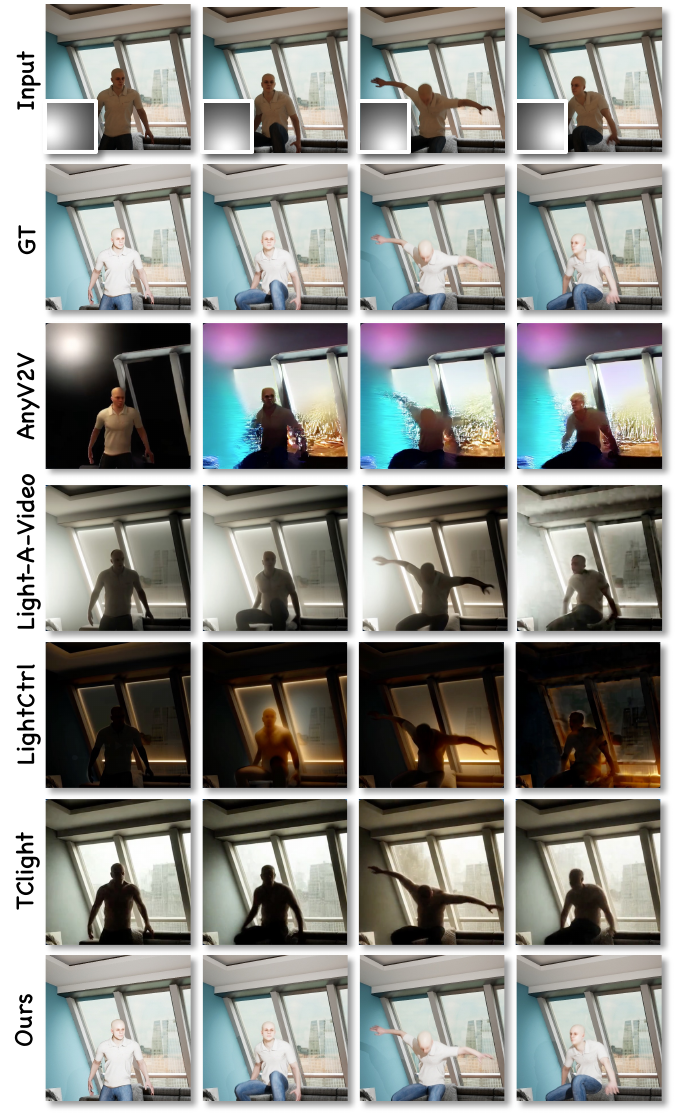}
  \caption{\textbf{Synthetic data evaluation.} We evaluate the performance of LiveLight on synthetic data to assess its ability to handle controlled lighting scenarios. LiveLight demonstrates superior performance in maintaining appearance consistency and temporal coherence under various lighting conditions.}
  \label{fig:synthetic_data}
\end{figure}

\begin{figure}[t]
  \centering
  \includegraphics[width=0.95\linewidth]{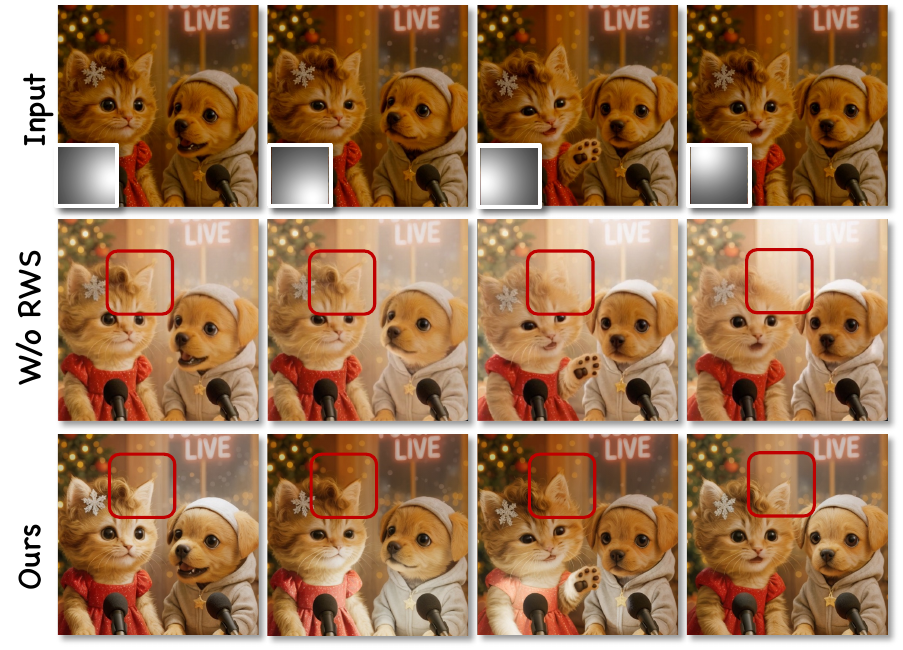}
  \caption{\textbf{Visual ablation study of the rolling-window strategy.}
Removing the rolling-window strategy leads to temporal flickering and unstable lighting across chunks. By propagating through the rolling window, our method maintains coherent illumination and stable appearance over time.}
  \label{fig:ablation_rws}
\end{figure}

\begin{figure}[t]
  \centering
  \includegraphics[width=0.95\linewidth]{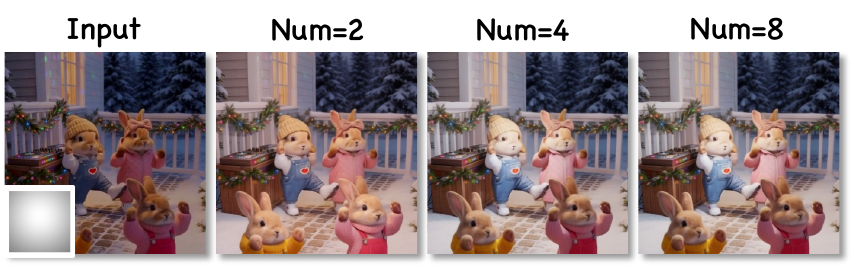}
  \caption{\textbf{Visual ablation study of the number of MPLI planes.} With $P{=}2$, the model cannot capture depth-dependent lighting, limiting 3D controllability. $P{=}8$ yields marginal gains over $P{=}4$ with added parameters. We adopt $P{=}4$ as the default.}
  \label{fig:ablation_plane}
\end{figure}

\begin{figure}[t]
  \centering
  \includegraphics[width=0.95\linewidth]{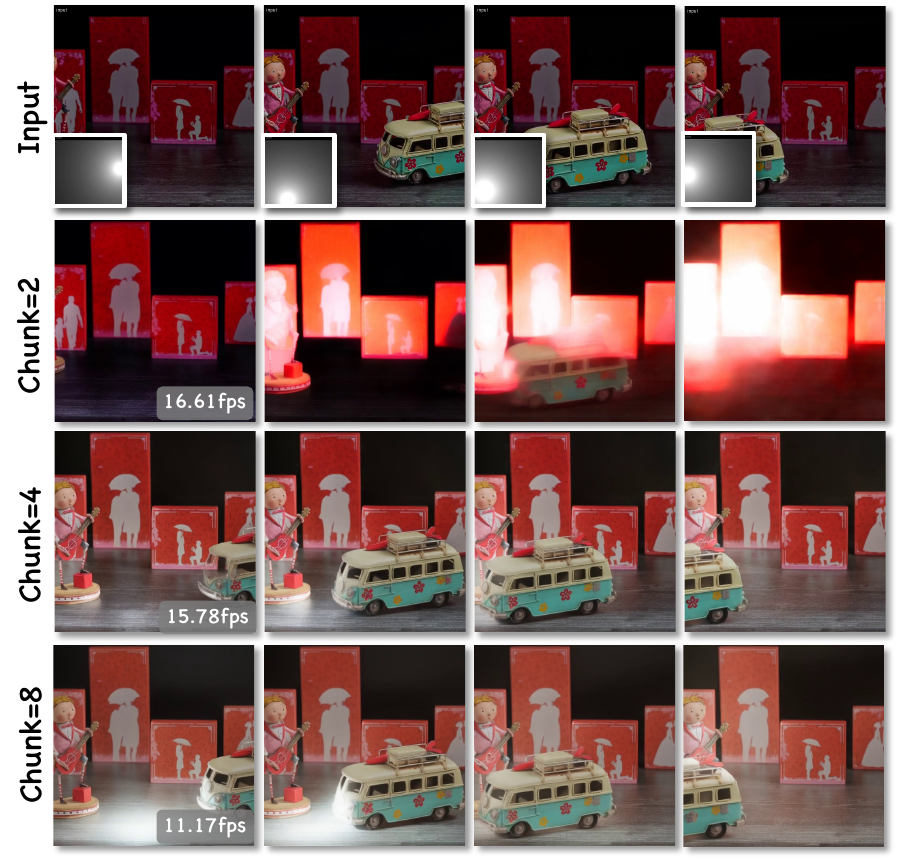}
  \caption{\textbf{Visual ablation study of micro-chunk size.} $M{=}2$ (16.61\,fps) causes severe temporal instability. $M{=}8$ (11.17\,fps) improves coherence but reduces throughput. We adopt $M{=}4$ (15.78\,fps) for the best latency--quality balance.}
  \label{fig:ablation_chunk}
\end{figure}

\subsection{Ablation Study}

To validate each design choice in LiveLight, we systematically ablate key components spanning all three training stages. Table~\ref{tab:ablation_component} reports the results of replacing or removing individual modules, and Table~\ref{tab:ablation_numerical} compares different numerical hyperparameter settings. All variants are evaluated on the same synthetic test set with identical inference settings unless otherwise stated.

\begin{figure}[t]
  \centering
  \includegraphics[width=0.95\linewidth]{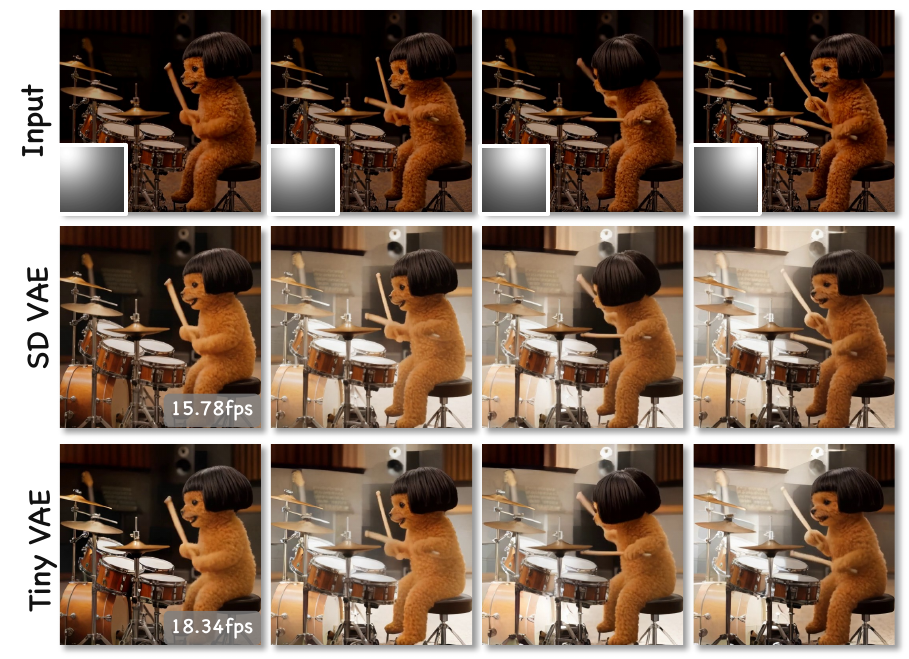}
  \caption{\textbf{Visual ablation study of VAE decoder.} The Tiny VAE (18.34\,fps) produces visual quality comparable to the SD VAE (15.78\,fps) while offering faster decoding, making it well suited for real-time streaming.} 
  \label{fig:ablation_vae}
\end{figure}

\noindent{Effectiveness of VAE decoder.}
Our framework is compatible with lightweight VAE decoders. As shown in Fig.~\ref{fig:ablation_vae} and Table~\ref{tab:ablation_component}, replacing the standard SD VAE with Tiny VAE~\citep{taesd_hf2023} increases the throughput from 15.78\,fps to 18.34\,fps with only a marginal quality drop, demonstrating that the VAE decoder is not the quality bottleneck and can be swapped for faster alternatives when latency is prioritized. In all other experiments, we use the standard SD VAE for fair comparison with baselines.

\begin{figure}[t]
  \centering
  \includegraphics[width=0.95\linewidth]{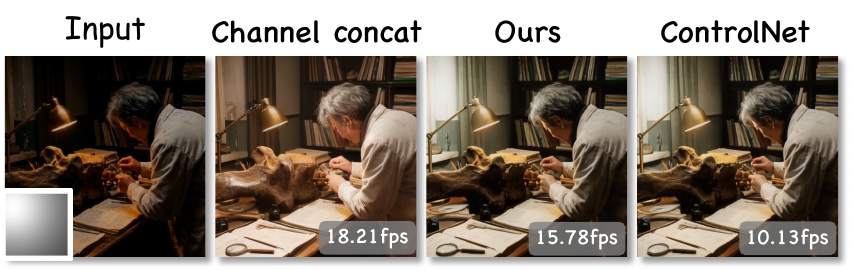}
  \caption{\textbf{Visual ablation study of light injection strategy.} Channel concatenation (18.21\,fps) disrupts the pretrained prior, degrading quality. ControlNet (10.13\,fps) preserves the prior but nearly halves throughput. Our adapter (15.78\,fps) achieves the best quality--efficiency trade-off.}
  \label{fig:ablation_lightadapter}
\end{figure}

\noindent{Effectiveness of light injection strategy.}
We compare three strategies for injecting MPLI lighting conditions in Fig.~\ref{fig:ablation_lightadapter}: channel concatenation (18.21\,fps), our lightweight adapter (15.78\,fps), and ControlNet (10.13\,fps). Channel concatenation modifies the input channels and requires retraining the backbone from scratch, disrupting the pretrained generative prior and degrading output quality. ControlNet preserves the prior but nearly halves the throughput due to its duplicate encoder. Our adapter retains the full pretrained capability with only a modest speed cost, achieving the best quality--efficiency trade-off.

\noindent{Effectiveness of Number of MPLI planes.}
We vary the number of depth planes $P \in \{2, 4, 8\}$ in Fig.~\ref{fig:ablation_plane}. With only 2 planes, the model struggles to capture depth-dependent lighting variation, limiting 3D lighting controllability. Increasing to 8 planes yields marginal improvement over 4 planes while adding extra parameters. We adopt $P{=}4$ as the default, balancing 3D control capability and model compactness.

\begin{figure}[t]
  \centering
  \includegraphics[width=0.95\linewidth]{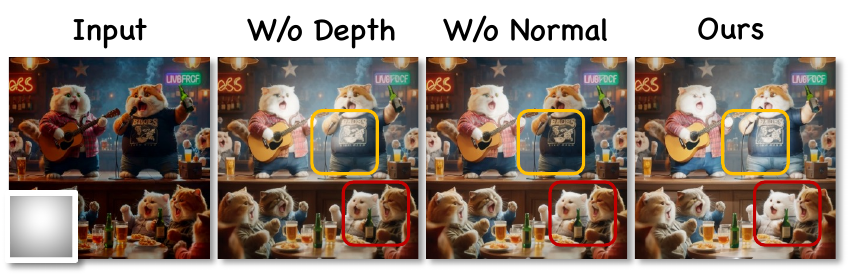}
  \caption{\textbf{Visual ablation study of the geometry-guided feedback branch.} Removing depth (W/o Depth) or normal (W/o Normal) supervision produces highlights and shadows inconsistent with the surface geometry. The full feedback yields the most geometrically plausible relighting.}
  \label{fig:ablation_feedback}
\end{figure}

\noindent{Effectiveness of geometry constraints.}
We ablate the geometry-aware feedback branch in Fig.~\ref{fig:ablation_feedback} by removing depth supervision (W/o Depth) or normal supervision (W/o Normal). Removing either component leads to less accurate highlights and shadows that are inconsistent with the underlying surface geometry. The full feedback configuration produces the most geometrically plausible relighting results.

\begin{figure}[t]
  \centering
  \includegraphics[width=0.95\linewidth]{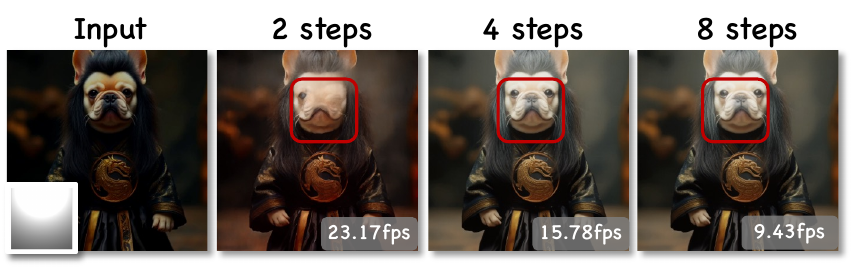}
  \caption{\textbf{Visual ablation study of distillation denoising steps.} Distilling to 2 steps (23.17\,fps) introduces visible facial artifacts. Using 8 steps (9.43\,fps) yields marginal quality gains over 4 steps (15.78\,fps). We adopt 4 steps as the default for optimal quality--speed balance.}
  \label{fig:ablation_step}
\end{figure}

\noindent{Effectiveness of distillation steps.}
We compare 2, 4, and 8 denoising steps in Fig.~\ref{fig:ablation_step}. Aggressive distillation to 2 steps runs at 23.17\,fps but introduces visible artifacts in details. Using 8 steps improves quality marginally over 4 steps while reducing throughput to 9.43\,fps. We adopt 4 steps (15.78\,fps) as the default, achieving a favorable balance between fidelity and real-time performance.

\noindent{Effectiveness of micro-chunk size.}
We study the effect of chunk size $M \in \{2, 4, 8\}$ on temporal quality and inference speed. As shown in Fig.~\ref{fig:ablation_chunk} and Table~\ref{tab:ablation_numerical}, a smaller chunk size $M{=}2$ yields the fastest throughput (16.61\,fps) but suffers from noticeable temporal instability due to insufficient inter-frame context within each chunk. Conversely, a larger chunk size $M{=}8$ provides stronger temporal coherence but significantly increases computation, reducing throughput to 11.17\,fps. We adopt $M{=}4$ (15.78\,fps) as the default, which achieves a favorable trade-off between temporal stability and real-time performance.

\noindent{Effectiveness of rolling-window strategy(RWS) .}
We ablate the rolling-window training strategy, which simulates the rolling-window inference process during training. As shown in Fig.~\ref{fig:ablation_rws} and Table~\ref{tab:ablation_component}, removing RWS leads to noticeable appearance drift and temporal inconsistency over long sequences, as the model has never been exposed to the error accumulation pattern inherent in streaming inference. With RWS, the model learns to maintain stable appearance and background details across frames, effectively reducing error accumulation and improving temporal coherence.

\begin{table}[t]
  \centering
  \caption{\textbf{Ablation study on component design choices.}
  Each row removes or replaces one component from the full model.
  \textcolor{Red}{\textbf{Red}} and \textcolor{Blue}{\textbf{Blue}} denote the best and second best results, respectively.}
  \label{tab:ablation_component}
  \resizebox{\linewidth}{!}{%
    \begin{tabular}{l|cccc}
      \toprule
      Variant
        & PSNR $\uparrow$
        & SSIM $\uparrow$
        & LPIPS $\downarrow$
        & Time (s) $\downarrow$ \\
      \midrule
      \multicolumn{5}{l}{\textcolor{gray}{\textit{VAE Decoder}}} \\
      \midrule
      Tiny VAE
        & 42.917 & 0.9762 & 0.1098 & \textcolor{Red}{\textbf{0.87}} \\
      \midrule
      \multicolumn{5}{l}{\textcolor{gray}{\textit{Light Injection}}} \\
      \midrule
      Channel concat
        & 39.843 & 0.9524 & 0.1536 & 0.88 \\
      ControlNet
        & \textcolor{Blue}{\textbf{43.108}} & \textcolor{Blue}{\textbf{0.9763}} & \textcolor{Blue}{\textbf{0.1082}} & 1.58 \\
      \midrule
      \multicolumn{5}{l}{\textcolor{gray}{\textit{Geometry-guided Feedback}}} \\
      \midrule
      W/o Depth Feedback
        & 41.726 & 0.9681 & 0.1274 & 1.01 \\
      W/o Normal Feedback
        & 42.183 & 0.9714 & 0.1193 & 1.01 \\
      \midrule
      \multicolumn{5}{l}{\textcolor{gray}{\textit{Rolling window streaming Training}}} \\
      \midrule
      W/o RWS
        & 41.392 & 0.9653 & 0.1342 & 1.01 \\
      \midrule
      LiveLight (Ours)
        & \textcolor{Red}{\textbf{43.264}} & \textcolor{Red}{\textbf{0.9778}} & \textcolor{Red}{\textbf{0.1064}} & 1.01 \\
      \bottomrule
    \end{tabular}%
  }
\end{table}

\begin{table}[t]
  \centering
  \caption{\textbf{Ablation study on numerical hyperparameters.}
  \textcolor{Red}{\textbf{Red}} and \textcolor{Blue}{\textbf{Blue}} denote the best and second best results, respectively.}
  \label{tab:ablation_numerical}
  \resizebox{\linewidth}{!}{%
    \begin{tabular}{l|cccc}
      \toprule
      Variant
        & PSNR $\uparrow$
        & SSIM $\uparrow$
        & LPIPS $\downarrow$
        & Time (s) $\downarrow$ \\
      \midrule
      \multicolumn{5}{l}{\textcolor{gray}{\textit{MPLI Planes ($P$)}}} \\
      \midrule
      $P=2$
        & 40.538 & 0.9586 & 0.1387 & 1.01 \\
      $P=8$
        & \textcolor{Blue}{\textbf{43.198}} & \textcolor{Blue}{\textbf{0.9771}} & \textcolor{Blue}{\textbf{0.1072}} & 1.06 \\
      \midrule
      \multicolumn{5}{l}{\textcolor{gray}{\textit{Distillation Steps}}} \\
      \midrule
      Step $=2$
        & 39.274 & 0.9483 & 0.1592 & \textcolor{Red}{\textbf{0.69}} \\
      Step $=8$
        & 43.182 & 0.9768 & 0.1076 & 1.70 \\
      \midrule
      \multicolumn{5}{l}{\textcolor{gray}{\textit{Chunk Size ($M$)}}} \\
      \midrule
      $M=2$
        & 40.847 & 0.9612 & 0.1356 & \textcolor{Blue}{\textbf{0.96}} \\
      $M=8$
        & 43.156 & 0.9765 & 0.1079 & 1.43 \\
      \midrule
      LiveLight ($P{=}4$, Step${=}4$, $M{=}4$)
        & \textcolor{Red}{\textbf{43.264}} & \textcolor{Red}{\textbf{0.9778}} & \textcolor{Red}{\textbf{0.1064}} & 1.01 \\
      \bottomrule
    \end{tabular}%
  }
\end{table}

\begin{figure}[t]
  \centering
  \includegraphics[width=0.95\linewidth]{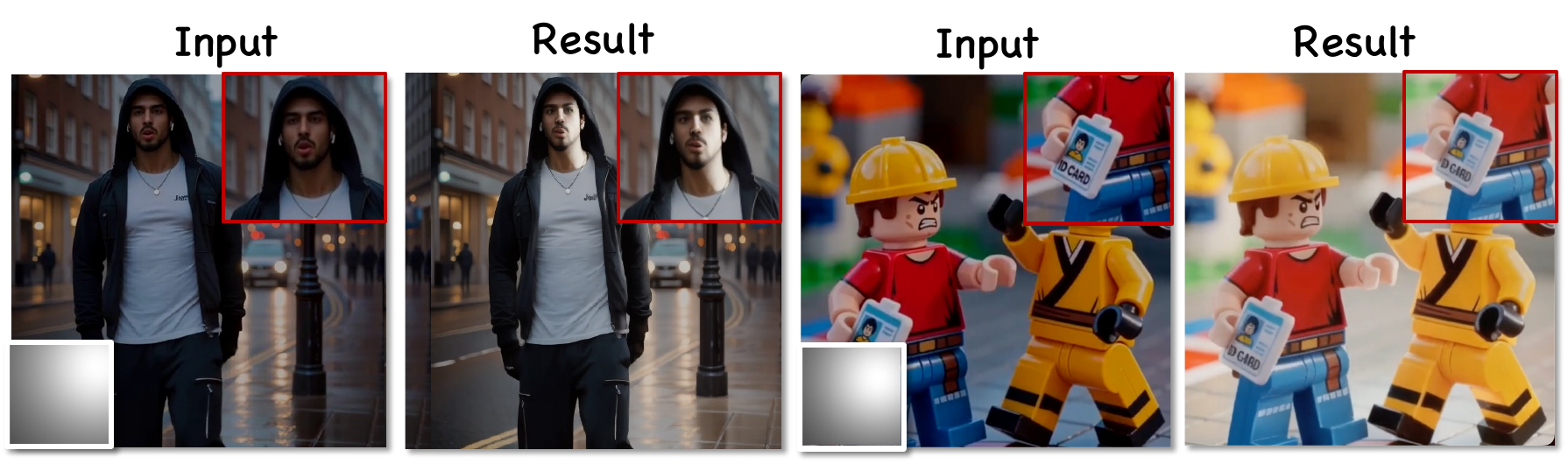}
  \caption{\textbf{Failure cases.} Left: a distant face loses fine facial details after relighting. Right: small text (``ID CARD'') and accessories on Lego figures exhibit visible artifacts. These failures are primarily attributed to the limited reconstruction capacity of the SD VAE decoder for high-frequency details.}
  \label{fig:failure}
\end{figure}

\section{Limitation}

Despite the promising results, LiveLight has several limitations, as illustrated in Fig.~\ref{fig:failure}. First, our method struggles with small or distant faces, where the limited spatial resolution leads to degraded relighting quality and loss of fine facial details. Second, we observe occasional artifacts in high-frequency details such as text and small accessories in the generated frames. These two issues are primarily attributed to the decoder capacity of the Stable Diffusion VAE~\cite{rombach2022high}, which was not designed to faithfully reconstruct such fine-grained details. We expect that migrating to a stronger base model with a more capable decoder can alleviate this issue. Finally, the geometry-guided feedback branch relies on the accuracy of off-the-shelf depth and normal estimators. When the geometry estimation is inaccurate—particularly for complex hairstyles, occlusions, or extreme poses—the shading guidance may be suboptimal, leading to less physically plausible relighting.  

\section{Conclusion}

We present LiveLight, the first diffusion-based framework for real-time, streaming video relighting with interactive 3D lighting control. Unlike existing offline methods that require the full lighting trajectory before generation, LiveLight allows users to dynamically adjust light position, intensity, and color while the relit video is produced frame by frame with low latency. Our framework combines a lightweight MPLI adapter for spatially precise and depth-aware illumination control, a geometry-guided feedback branch for geometry-consistent shading, and a progressive micro-chunk streaming strategy for temporally coherent, arbitrarily long video generation. Extensive experiments demonstrate that LiveLight achieves state-of-the-art quality while running at real-time speed, outperforming offline baselines in temporal stability, controllability, and user preference. We hope LiveLight inspires future research on real-time controllable video generation.

\section{Potential Societal Impacts}

LiveLight enables real-time, interactive relighting of portrait videos, which can benefit creative professionals in film production, virtual cinematography, and live-streaming content creation by significantly reducing the cost and time of lighting adjustments. However, as with other generative video techniques, the ability to realistically alter lighting in portrait videos could potentially be misused to create misleading or deceptive visual content. We encourage the research community to develop robust detection and watermarking mechanisms alongside generative methods, and to establish responsible usage guidelines. Our work focuses on lighting manipulation rather than identity or content generation, which inherently limits its misuse potential compared to general-purpose video synthesis methods.

\par\medskip
\noindent\textbf{Acknowledgments.}\enspace
This work was supported by HKUST under Grant No.~WEB25EG01.\par

{
    \small
    \bibliographystyle{ieeenat_fullname}
    \bibliography{main}
}

\end{document}